\documentclass[letterpaper, 10 pt, journal, twoside]{IEEEtran}

\usepackage{graphics} % for pdf, bitmapped graphics files
\usepackage{epsfig} % for postscript graphics files
\usepackage{newtxtext,newtxmath} % assumes new font selection scheme installed
\usepackage{times} % assumes new font selection scheme installed
\usepackage[font=scriptsize]{caption}
\usepackage{subcaption}
\usepackage{amsmath} % assumes amsmath package installed
\usepackage[utf8]{inputenc}
\usepackage[T1]{fontenc}
\usepackage{mathtools}
\usepackage{enumitem}
\usepackage{xcolor}
\usepackage{float}
\usepackage{tabularx}
\newcolumntype{Y}{>{\centering\arraybackslash}X}

 \usepackage{hyperref}
\usepackage{xcolor}
\usepackage{comment}

\usepackage{xcolor,colortbl}

\definecolor{Gray}{gray}{0.85}
\definecolor{LightCyan}{rgb}{0.88,1,1}

\newcolumntype{a}{>{\columncolor{Gray}}c}
\ifCLASSINFOpdf
\else
\fi
\begin{document}
\title{Dynamic Control of  Soft Robotic Arm: An Experimental Study}

\author{Milad~Azizkhani,
        Anthony L.~Gunderman,
        Isuru~S.~Godage, and Yue~Chen% <-this % stops a space
\thanks{
\noindent This research is supported by the Georgia Tech IRIM seed grant. \\
Milad Azizkhani and Anthony Gunderman are with Georgia Tech Institute for Robotics and Intelligent Machines (IRIM), Atlanta, GA 30313, USA ([mazizkhani3, agunderman3]@gatech.edu). 

Isuru S. Godage is with the Engineering Technology \& Industrial Distribution, Texas A\&M University,
College Station, TX   77843, USA (igodage@exchange.tamu.edu)

Yue Chen  is with   Georgia Tech Institute for Robotics and Intelligent Machines (IRIM) and the School of Biomedical Engineering, Georgia Institue of Technology, Atlanta, GA 30313, USA (yue.chen@bme.gatech.edu).

This work has been submitted to the IEEE for possible publication. Copyright may be transferred without notice, after which this version may no longer be accessible.

}}

\maketitle

\begin{abstract}
In this paper, a reinforced soft robot prototype with a custom-designed actuator-space string encoder are created to investigate dynamic soft robotic trajectory tracking. The soft robot prototype embedded with the proposed adaptive passivity control and efficient dynamic model  make the challenging trajectory tracking tasks  possible. We focus on the exploration  of  tracking accuracy as well as the full potential of the proposed control strategy by performing  experimental validations at different operation scenarios: various tracking speed and  external disturbance.  
In all experimental scenarios, the proposed adaptive passivity control outperforms the conventional PD feedback linearization control. The experimental analysis details the advantage and shortcoming of the proposed approach, and points out the next steps for future soft robot dynamic control.

\end{abstract}


% \fi
\IEEEpeerreviewmaketitle

\vspace{-4 mm}
\section{Introduction}
\IEEEPARstart{S}{o}ft robots, inspired from nature, have been introduced to the robotics community to provide abilities that can surpass their rigid counterparts \cite{laschi2016soft}. It has been observed in nature that biological animals or plants usually present more adaptive, agile, and versatile behavior with respect to man-made machines. Soft robots are made from inherently elastic materials such as silicone or rubber, which makes the robot safe and compliant. Theoretically, soft robots can be considered as infinite degree of freedom robots since elastic materials can present different motions ranging from elongation, twisting, bending, etc %due to their continuum nature
\cite{rus2015design}. These behavior made soft robots a promising tool for different application such as grasping \cite{amend2012positive, arachchige2021novel}, medical procedures \cite{li2021development, musa2022mri}, rehabilitation purposes \cite{pan2022soft}, berry harvesting \cite{gunderman2022tendon}, snake locomotion \cite{D2},  etc.

Manipulation  has always been considered as one of the most important robotic applications. The elephant trunk  has shown great dexterity to manipulate and grasp objects with different shape and sizes \cite{dagenais2021elephants}. To mimic those animal behaviors,  different  manipulators have been introduced in literature such as Air-OCTOR \cite{jones2006kinematics}, modular hybrid continuum arm \cite{Yang}, OCT ARM V \cite{MC} and many more. Manipulation accuracy  and speed are the two primary performance metrics to evaluate these soft robotic arms. Despite many progress have been made in the past two decades, their real-time dynamic control capability is still underdeveloped. 

Generally, model-based control has been of great interest of robotics researchers due to the availability of stability proof  and readily understandable formulation. To obtain the desired results, having an accurate model could be very helpful to cancel out the dynamics and enforce desired behavior through controller. 
Accurate modeling  in soft robots such as cosserat rod theory \cite{till2019real} and finite element methods\cite{Poly} are challenging to be implemented in real-time control applications due to their computationally expensive calculations. However, it should be noted that several  groups are addressing these problems with interesting ideas such as  \cite{largilliere2015real, alqumsan2019robust}.

Lumped parameter models, derived from piece-wise constant curvature (PCC) assumption and Euler-Lagrange formulation,  can be considered as a suitable candidates for real-time implementation. However, it is obvious that the lumped parameter model will have some uncertainties and cannot accurately predict the robot dynamics in all situations. To resolve this issue, a sophisticated control algorithm that can address the unknown and partially unknown factors is required to achieve desired performance. 

Many works have investigated the    control strategies to compensate the unknown or partially known uncertainties in soft robot control. For instance, in a 1-DOF soft robot, 

Azizkhani et. al \cite{azizkhani2022model} considered the actuator dynamics as a second-order model, and the variation in system parameters along with unknown dynamics and external disturbances has been compensated with a robust model reference adaptive control with considering input constraints in the input pressure. Godage et. al \cite{godage2018dynamic} modeled the system dynamics as a second order system with hysteresis and used it to cancel out the dynamics. The remaining effects have been compensated by a PID controller. In another study \cite{xavier2022nonlinear}, a nonlinear control with unscented kalamn filter (UKF) with feedback linearization has been used to control a bending actuator. 

With promising results for 1-DoF system, now is the time to explore the control algorithm for the high-dimensional system. In our previous simulation study \cite{azizkhani2022dynamic}, we have shown that an adaptive passivity control with sigma modification can handle parametric uncertainty, external disturbances, payload, sensor noise, and fast tracking and achieve improved performance compared to the conventional PD + Feedback linearization   \cite{della2020model,falkenhahn2016dynamic}. Other studies have also focused on the adaptive controller to address the uncertainties and unknown dynamics in soft robotic arms such as implementing adaptive control \cite{trumic2021adaptive} inspired by \cite{slotine1987adaptive}, and adaptive terminal sliding mode control \cite{kazemipour2022adaptive} to improve convergence speed.  

In this paper, we are building upon our previous work \cite{azizkhani2022dynamic} to implement the proposed controller on a robust, reinforced soft robotic arm. The new design has focused on improving the payload handling with increased stiffness, and modular design. We try to address three problems in this paper: (1) incorporate a custom designed rotary encoder to eliminate the need for an observer-based velocity measurements in actuator space, (2) investigate the dynamic tracking speed and accuracy of the proposed controller, and (3) compensate the  disturbance and uncertainties during trajectory tracking. Despite numerous potential for soft robotic manipulators, there is still an obvious gap in fast and accurate robot tracking. Therefore, the purpose of this study is to investigate the potential and limits of the proposed   controller to address the limitations  of current approach and pave the road to the versatile soft robot manipulation as what the rigid robotic arms  have brought over the past decades.

The rest of the paper is organized as follows. In section \ref{sec:Hardware Design}, the modular soft robotic arm design, fabrication, and system integration is discussed. Section \ref{sec:System modeling} will briefly review the kinematic and dynamic model and discuss how the new design  will contribute to model uncertainty. Section \ref{sec:Control Design} discuss controllers derivation based on PD feedback linearizaton and adaptive passivity. In section \ref{experimentalstudy}, the system experimental result and discussions are provided. Finally, the conclusion of this paper is presented in section \ref{sec:Conclusion}.

\begin{figure}[t]
    \centering
    \includegraphics[width=0.4\textwidth]{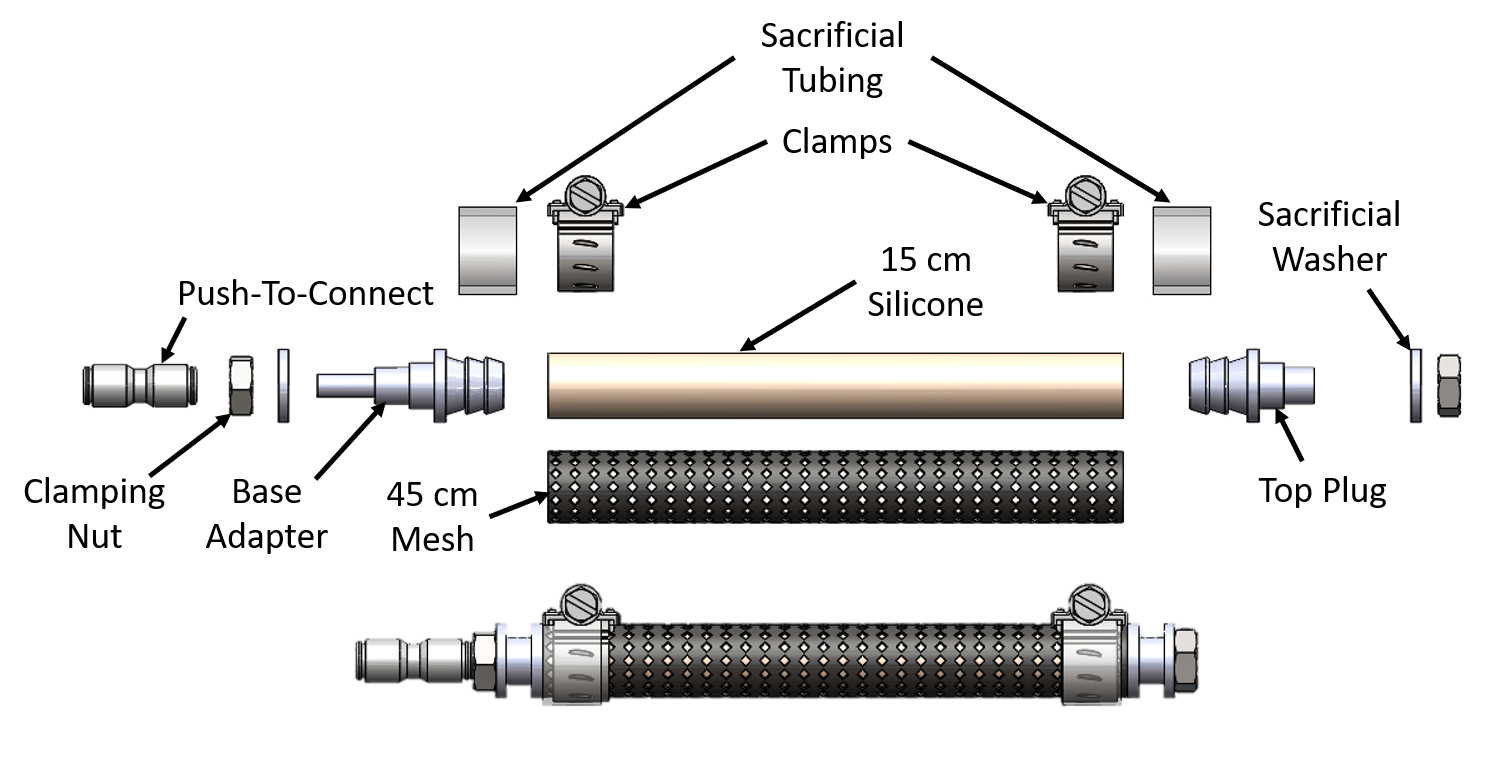}
    \caption{Exploded view of modular soft robotic actuator}
    \label{fig:ActutorDesign}
    \vspace{-6 mm}
\end{figure}
\vspace{-5mm}
\section{Hardware Design}
\label{sec:Hardware Design}

The robotic hardware is expected to achieve two basic features. First, the system is fabricated in a way that minimizes the tearing down of the elastic material to increase the life cycle. Second, 
the system is designed in a modular fashion so that the replacement process is easy and fast in the scenario when  one of the actuators fails.
\vspace{-3 mm}
\subsection{Actuator Design}
The elongation actuator is consisted of 15 cm silicone rubber  and 45 cm covering mesh. Two plugs for the distal and proximal end have been designed to provide a good fitting for the actuator and the modularity to be attached to the base of the soft robot section. Two clamps are used at both ends to prevent any leaking of air. Two sacrificial tubings will be used to mitigate the tearing forces acting on the base of silicone rubber. Then, clamping nuts and washers are used to connect the actuator to the main plates of the soft robot section. The exploded view of the described design is detailed in Fig \ref{fig:ActutorDesign}.

\vspace{-5 mm}
\subsection{Section Design}
\begin{figure}
    \centering
    \includegraphics[width=0.4\textwidth]{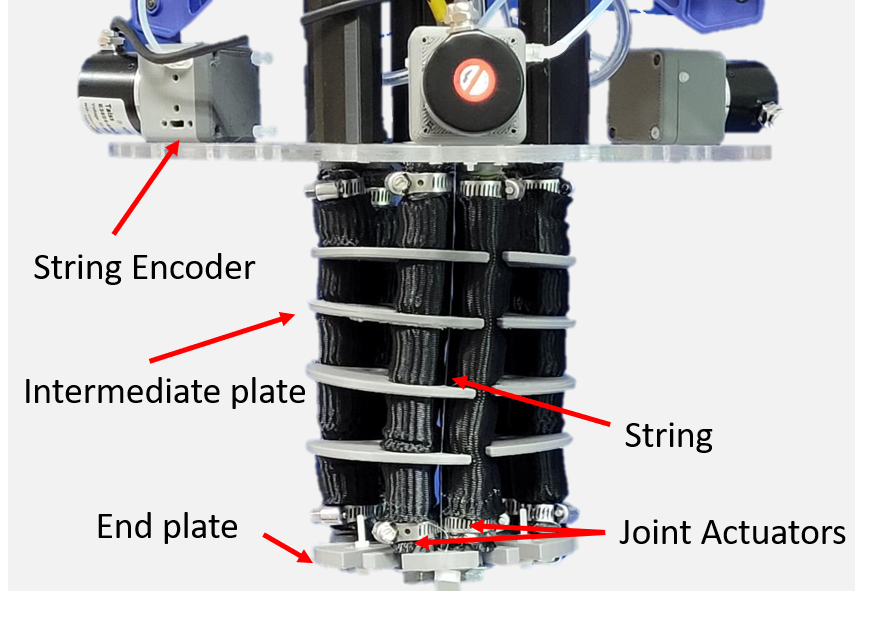}
    \caption{3-DoF soft robot prototype along with the proposed string encoders for actuator-space measurement.}
    \label{fig:SectionDesign}
    \vspace{-6 mm}
\end{figure}

As discussed in \cite{azizkhani2022dynamic}, the placement location of actuators is 120 degrees apart from each other with an equal distance from the center. To enhance the robot stiffness for high-demand applications, such as picking up an object,  two actuators have been placed in parallel to create the reinforced  elongation actuator. Note that their input lets are connected via a T-shape fitting and they are actuating at the same time. As can be seen in Fig. \ref{fig:SectionDesign}, 4 intermediate planes have been used to constrain the robot to follow a piece-wise constant curvature (PCC) assumption.
\vspace{-3 mm}
\subsection{String Encoder Design}
\begin{figure}
    \centering
    \includegraphics[width=0.35\textwidth]{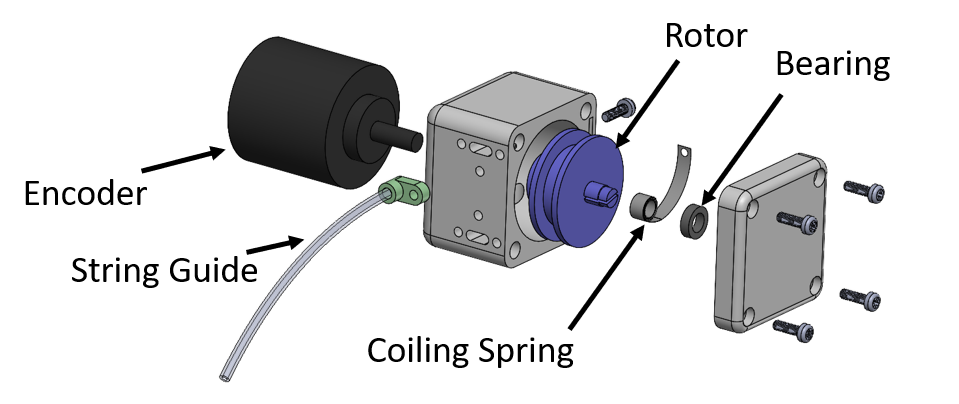}
    \caption{Exploded view of string encoder.}

    \label{fig:Encoder}
    \vspace{-6 mm}
\end{figure}

The kinematic variables of a soft robotic arm are the length change of each actuator. Therefore, designing a sensor that can measure the elongation of the actuator would be of great importance. According to Fig. \ref{fig:SectionDesign}, a string (fishing line) has been routed through holes in all the intermediate, proximal and distal plates with a fixed offset from actuators. By using a simple linear ratio, the offset can be compensated and true elongation can be obtained. The string is wrapped around a commercialized encoder rotor which translates the elongation to a rotary motion.
It should be noted that the fishing line should always be pre-loaded to return a correct value. To ensure tension along the fishing line, a coiling spring has been attached to the rotor and the end of the shaft is placed into a bearing for smooth rotation. The exploding view of the proposed rotary encoder is shown in Fig. \ref{fig:Encoder}. The fabricated sensor function is similar to off-the-shelf string potentiometers with the primary advantage of being cost-effective and high resolution. With basic geometry knowledge, the relation between elongation and rotation is described as follows: 
\begin{equation}
    \Delta l = \frac{2\pi r_e}{4n} p
\end{equation}
where $r_e$ is the radius of the string encoder rotor, $n$ is the number of pulses per revolution, 4 is for quadrature reading of the encoder, and $p$ is the input pulse.

\vspace{-2 mm}
\section{System modeling}
\label{sec:System modeling}
\subsection{Kinematic Modeling}
In this section, we briefly review  kinematic modeling of the soft robotic arm based on \cite{godage2015modal,azizkhani2022dynamic}. To define the robot's position in 3D space, two mappings have been used: (1) actuator space to configuration space, and (2) configuration space to task space.
\subsubsection{Mapping from actuator space to configuration space}
To define the relationship between actuator variables to configuration variables, the length of actuators has been considered as follows
\begin{equation}
    L_{i}(t) = L_{0} + l_{i}(t)
\end{equation}
where $L_{i}$ represent $i^{th}$ actuator total length, $L_0$ is the initial length, and $l_{i}$ is the $i^{th}$ actuator elongation. The actuator variables vector is represented by $q = [l_{1},l_{2},l_{3}]$  and the mapping is derived based on the geometrical location of actuators (120 degrees apart with a constant radius from the center of the base) and PCC assumption as follows: 
\begin{equation}
    \begin{aligned}
            s &= \sqrt{{l_{1}}^2+{l_{2}}^2 + {l_{3}}^2 -l_{1}l_{2} - l_{2}l_{3} - l_{1}l_{3}}, \   \phi = \frac{2s}{3r}\\
            \lambda &= \frac{(3L_{0}+l_{1}+l_{2}+l_{3})r}{2s}, \ \theta = \arctan\left({\frac{\sqrt{3}(l_{3}-l_{2})}{l_{2} + l_{3} - 2l_{1}}}\right)
    \end{aligned}
\end{equation}
where $\phi \in [0, 2\pi)$ is the bending angle of the arc, $\lambda \in (0, \infty)$ is the radius of the curvature, and $\theta \in (-\pi, \pi)$ is the bending angle with respect to X-Axis. 
\subsubsection{Mapping From Configuration Space to Task Space}
The homogeneous transformation $T \in SE(3)$ is defined as follows:  
\begin{equation}
    T(\xi,q) = \text{Rot}_{z}(\theta)\text{Trans}_{x}(\lambda)\text{Rot}_{y}(\xi \phi) \text{Trans}_{x}(-\lambda)\text{Rot}_{z}(-\theta)
\end{equation}
where $\text{Rot}_{i}$ defines homogeneous transformation in $SE(3)$ for rotation around $i$ axis and ${Trans}_{i}$ translation along $i$ axis. $\xi \in [0,1]$ is an auxiliary variable that defines the imaginary disk between proximal $\xi = 0$ to distal end $\xi =1$. 
\subsection{Dynamic Modeling}
The dynamic model of the robot, similar to our previous work \cite{azizkhani2022dynamic, godage2016dynamics}, can be derived using Euler-Lagrange formulation 
\begin{equation}
    M\boldsymbol{\Ddot{q}}+C\boldsymbol{\Dot{q}}+
    D\boldsymbol{\Dot{q}} + K\boldsymbol{q} + \boldsymbol{G} + \boldsymbol{H} = J^{T}\boldsymbol{F_{ext}} + \boldsymbol{\tau}
    \label{eq: dynamic-compact}
\end{equation}
where $M$ represents the inertia matrix, $C$ centrifugal and Coriolis forces, $D$ is the damping forces, $K$ the stiffness of the system, $G$ is the gravitational forces and $H$ represents hysteresis behavior. $F_{ext}$ represents external forces, $J$ is the jacobian of the system with respect to the base frame, and $\tau$ is the resulting forces from input pressure acting upon the actuators.

The inertia matrix is calculated as
\begin{equation}
    M = M^{w} + M^{v}
    \label{eq: M-matrix}
\end{equation}
where $M^{w}$ and $M^{v}$ are generalized angular and linear inertia matrices and are derived as follows
\begin{equation}
\begin{aligned}
            M^{w}[j,k] &= I_{xx} \int_{\xi} \mathbb{T}_{2} (\frac{\partial R}{\partial q(j)})(\frac{\partial R}{\partial q(k)})\\
            M^{v}[j,k] &= m \int_{\xi} (\frac{\partial \boldsymbol{P}}{\partial q(j)})(\frac{\partial \boldsymbol{P}}{\partial q(k)})
\end{aligned}
\end{equation}
where $R \in SO(3)$ represents the rotation matrix, and $\boldsymbol{P} \in R^{3}$ is the position of the robot, which is a function of $\zeta$ to describe all nodes from proximal to the distal end of the robot. 

The centrifugal and Coriolis matrix is defined as follows
\begin{equation}
    \begin{aligned}
    C[k,j] &= \sum_{i=1}^{3} \Gamma_{ijk}(M) \Dot{q}(i)\\
    \Gamma_{ijk}(M) &= \frac{1}{2} \left( \frac{\partial M[k,j]}{\partial q(i)} + \frac{\partial M[k,i]}{\partial q(j)}  -\frac{\partial M[i,j]}{\partial q(k)}\right)
    \end{aligned}
    \label{eq: C-Matrix}
\end{equation}
Furthermore, the gravitational forces are derived as
\begin{equation}
    G(i) = m \int_{\xi} {\boldsymbol{J^{v}}}^{T}(i)R^{T}\boldsymbol{G_{v}}
\end{equation}
where $J^{v}$ is the jacobian of the moving frame and is calculated as
\begin{equation}
    \boldsymbol{J^{v}}(i) = R^{T} \frac{\partial \boldsymbol{P}}{\partial q(i)}
\end{equation}
 
We would like to note that hysteresis behavior is an inherent characteristic of soft robot.  In this study, we will use the Bouc-Wen model \cite{godage2012pneumatic} to describe the hysteresis as follows
\begin{equation}
    \dot{h}(i) = q(i)\left[\alpha_{h} - \left\{\beta_{h} \text{sgn}(\dot{q}(i)h(i)) + \gamma_{h} \right\}|h(i)|  \right]
\end{equation}
where $\alpha_{h}$, $\beta_{h}$, $\gamma_{h}$ are the constant parameters of the model.

\vspace{-5 mm}
\section{Soft Robot Dynamic Controller Design}
\label{sec:Control Design}
\subsection{Benchmark Controller: PD Feedback Linearization (PDFL)}
In this section, we will first consider the widely used PD+Feedback linearizaition controller as the benchmark \cite{spong2008robot,azizkhani2022dynamic}, which is defined as follows: 
\begin{equation}
    \begin{aligned}
    \boldsymbol{\tau} &= \alpha \boldsymbol{\tau^{\prime}} + \boldsymbol{\beta},\quad \boldsymbol{\tau^{\prime}} = \boldsymbol{\Ddot{q}_{d}} - K_{d}\boldsymbol{\Dot{\tilde{q}}} - K_{p}\boldsymbol{{\tilde{q}}}\\
    \alpha &= M, \quad \boldsymbol{\beta} = C\boldsymbol{\Dot{q}} + D\boldsymbol{\Dot{q}} + K\boldsymbol{q} + \boldsymbol{G}
    \end{aligned}
    \label{eq:PDC}
\end{equation}
where $\beta$ represents the feedback term that compensates the known dynamics, and $\tau^{\prime}$ is the PD controller that forces the system to follow a reference trajectory with desired dynamic behavior. $K_p$ and $K_d$ are $R^{3 \times 3}$ positive definite diagonal matrices where their elements are equal to $k_p$ and $k_d$ respectively.
\subsection{Proposed Control: Adaptive Passivity (AP)  Control}

To compensate for the modeling uncertaineis and external disturbance, we will propose the adaptive passivity control here, as detailed below: 
\begin{equation}
\begin{aligned}
&\boldsymbol{v} =\boldsymbol{\dot{q}^{d}}-\Lambda \boldsymbol{\tilde{q}}, 
\boldsymbol{a} =\boldsymbol{\dot{v}}=\boldsymbol{\ddot{q}^{d}}-\Lambda \boldsymbol{\dot{\tilde{q}}},
\boldsymbol{r} =\boldsymbol{\dot{q}}-\boldsymbol{v}=\boldsymbol{\dot{\tilde{q}}}+\Lambda \boldsymbol{\tilde{q}}\\
&\boldsymbol{\tau}= M(\boldsymbol{q}) \boldsymbol{a}+ C(\boldsymbol{q}, \boldsymbol{\dot{q}}) \boldsymbol{v}+ G(\boldsymbol{q})-K_{G} \boldsymbol{r} + \hat{K} \boldsymbol{q} + \hat{D} \boldsymbol{v}
\end{aligned}
\label{eq:adaptCont}
\end{equation}
where $\hat{K}$ and $\hat{D}$ are uncertain and varying parameters, and $K_g$ and $\Lambda$ are $R^{3 \times 3}$ positive definite diagonal matrices where their elements are equal to $k_g$ and $\Lambda_{AP}$ respectively. The advantage of implementing an adaptive passivity control lies in the adaptive nature of the algorithm to compensate uncertainties by adapting to changing the behavior of the system's parameter and canceling out the unmodeled dynamics. In this study, due to the usage of two sets of actuators as a virtual reinforced actuator, we have some modeling uncertainty in the $M$, $C$, and $G$ matrix which will introduce more uncertainty into the system, however, we expect the system maintains its stability. By considering the linear in  parameter formulation of the system, we can rewrite equation \eqref{eq:adaptCont} as follows: 
\begin{equation}
    \boldsymbol{\tau}=Y(\boldsymbol{q}, \boldsymbol{\dot{q}}, \boldsymbol{a}, \boldsymbol{v}) \boldsymbol{\hat{\theta}_{p}}-K_{G} \boldsymbol{r}
    \label{eq:AdaptPass}
\end{equation}
where $Y$ denotes the regressor matrix and $\hat{\theta}_p$ are uncertain parameters that need to be adapted. Due to computational complexity, in this study, the $C$ matrix has been removed from the control rule, and the inaccurate $M$ and $G$ are used in the control rule without any adaptation. The only parameters that will be updated are stiffness and damping variables that have dominant effects on the system dynamics. The other terms are expected to be compensated by robust term $-K_{G}r$. The regressor matrix and adaptable parameters are described as follows
\begin{equation}
\begin{aligned}
    Y &= [\text{diag}([q(1),q(2),q(3)]),\text{diag}([v(1),v(2),v(3)])]\\
    \boldsymbol{\hat{\theta}_{p}} &= [\hat{K}_{1},\hat{K}_{2},\hat{K}_{3},\hat{D}_{1},\hat{D}_{2},\hat{D}_{3}]^{T},
\end{aligned}
\end{equation}

Following our previous paper \cite{azizkhani2022dynamic}, the adaptation rule is derived as 
\begin{equation}
    \boldsymbol{\dot{\hat{\theta}}_{p}}=-\Gamma^{-1} Y^{T}(\boldsymbol{q}, \boldsymbol{\dot{q}}, \boldsymbol{a}, \boldsymbol{v}) \boldsymbol{r}
\end{equation}
where  $\Gamma$ is defined as a $R^{6 \times 6}$ positive definite diagonal matrix as follows: 
\begin{equation}
    \Gamma = \begin{bmatrix}
    \Gamma_K & 0 & 0 & 0 & 0 & 0 \\
    0 & \Gamma_K & 0 & 0 & 0 & 0 \\
    0 & 0 & \Gamma_k & 0 & 0 & 0 \\
    0 & 0 & 0 & \Gamma_D & 0 & 0 \\
    0 & 0 & 0 & 0 & \Gamma_D & 0 \\ 
    0 & 0 & 0 & 0 & 0 & \Gamma_D \\
        \end{bmatrix}
\end{equation}
where ${\Gamma_K}$ and ${\Gamma_D}$ are adaptation gains for stiffness and damping matrix respectively.

\vspace{-3 mm}
\section{Experimental Study}
\label{experimentalstudy}
\subsection{Experimental Setup}
The experimental setup consists of three proportional pressure regulators (ITV 1031-21N2BL4, SMC Corporation) with embedded pressure sensors for pressurizing the actuators. The valves are connected to a compressor with the output pressure of 70 psi. The encoders (LPD3806-600BM-G5-24C) have a two-phase quadrature output with 600 pulses per revolution. The control algorithm was implemented on Simulink xPC target machine that is connected to an analog output board (DAC6703, National Instrument) which provides 0-5 volt to pressure regulators. The encoders' output was measured with a 32-bit counter PCI (CNT32-8M, Contec). The algorithm, analog outputs, and measurements have been executed with a 1KHz sampling rate. For Kinematic validation, an optical camera (MicronTracker, Claronav, Toronto, Canada) was used which operates with 16 Hz.
\vspace{-3 mm}
\subsection{System Parameter Identification}
\begin{figure}[bth]
     \centering
     \begin{subfigure}[c]{0.24\textwidth}
     \includegraphics[width=\textwidth]{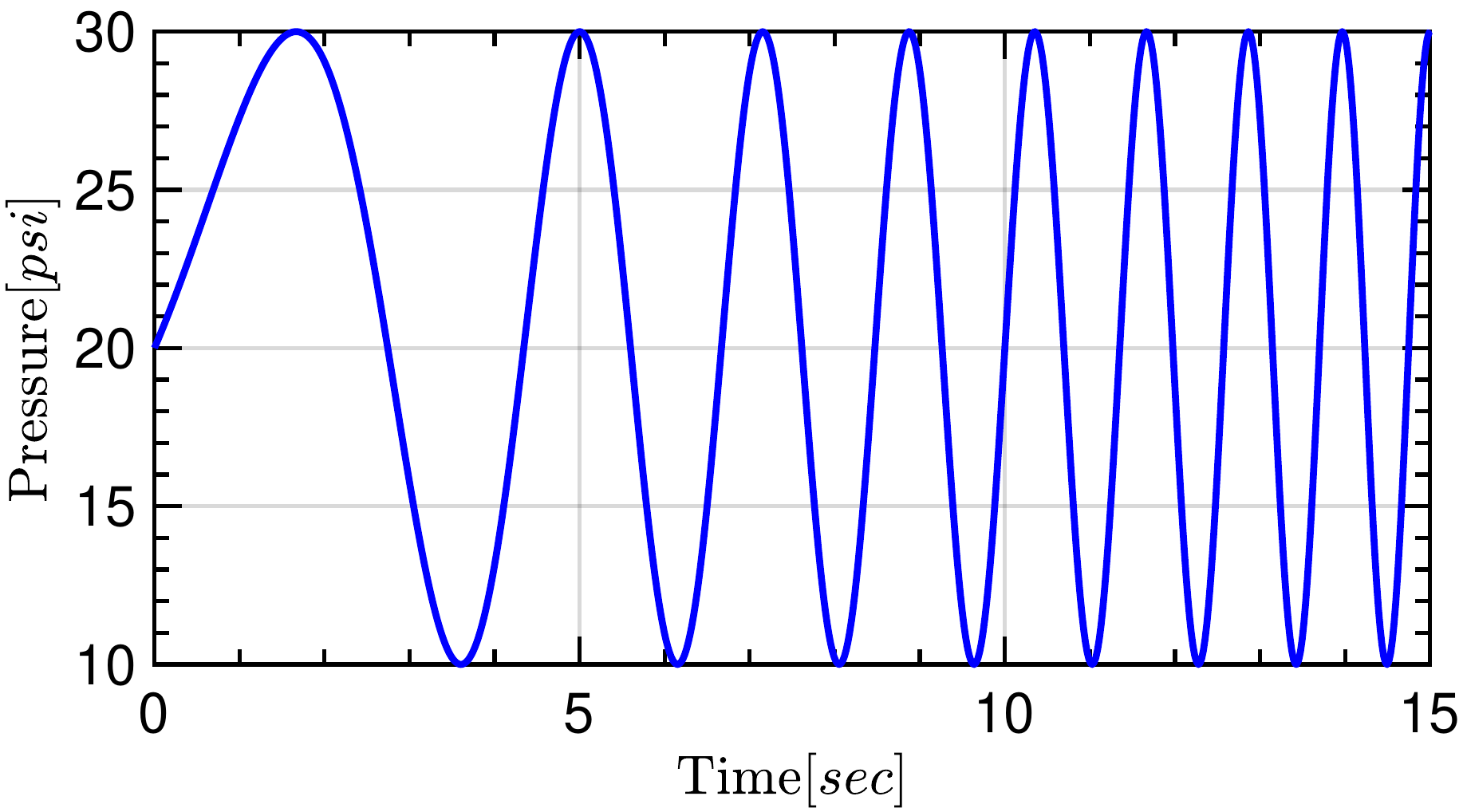}
     \end{subfigure}
     \begin{subfigure}[c]{0.24\textwidth}
     \includegraphics[width=\textwidth]{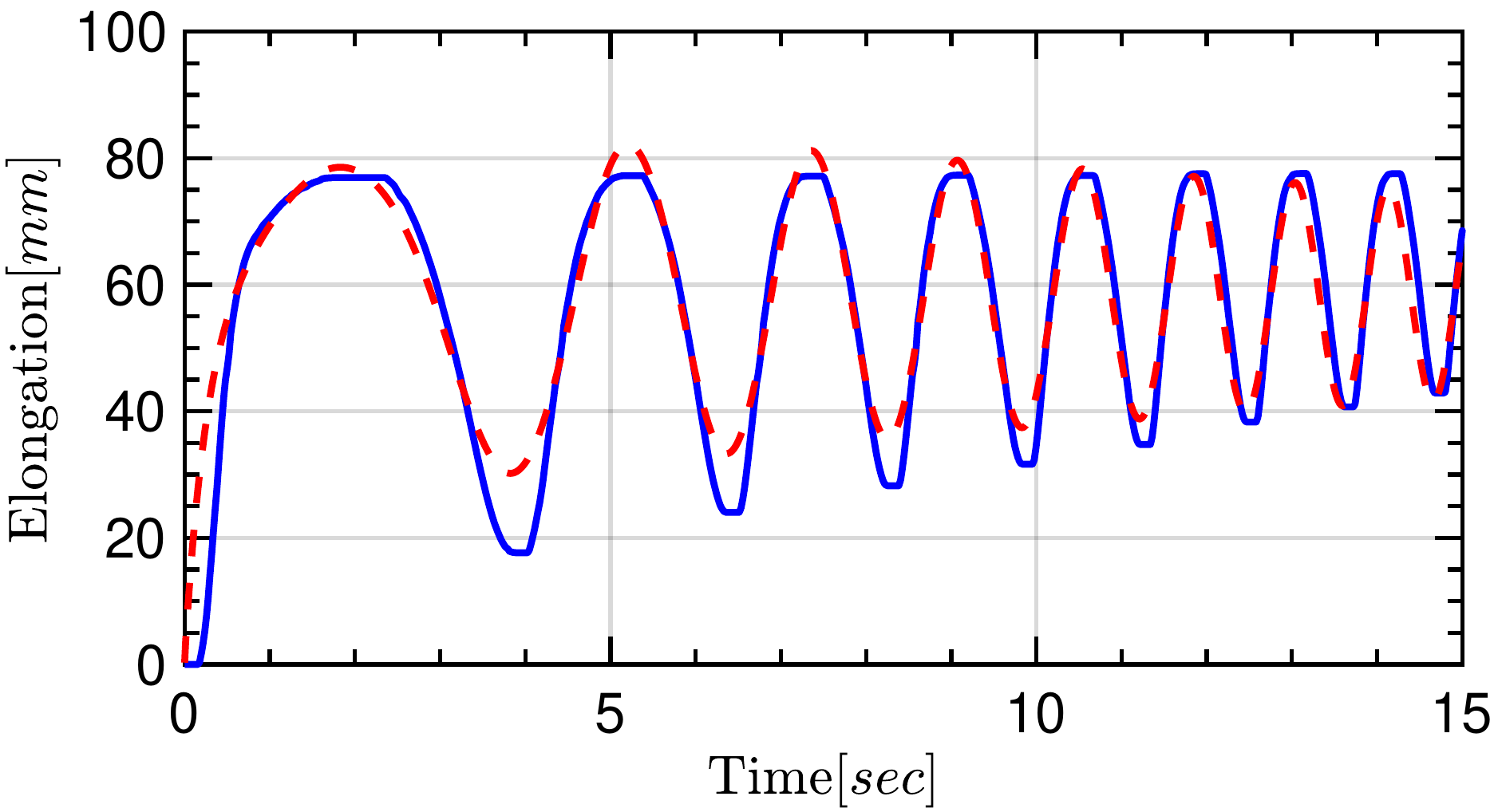}
     \end{subfigure}
     \caption{The left image depicts the input pressure sent to the system. In the right image, the blue shows the measured elongation from the robot and the red is the output from the simulated model in MATLAB.}
     \label{fig:IdentificationChirp}
     \vspace{-0.1in}
\end{figure}

To identify the system dynamic parameters, a
chirp signal was sent to one of the actuators to identify the system parameters. Noting that the parameters among these three actuators may vary due to imperfect manufacturing, but we expect the adaptive control can compensate for these uncertainties. 
The optimization algorithm is set to constraint nonlinear least square with cost function of sum squared error and is implemented in MATLAB. The input pressure and system response in simulation and robot hardware are depicted in Fig. \ref{fig:IdentificationChirp}.

The identified parameters are summarized as follows
\begin{equation}
\begin{aligned}
    D &= \begin{bmatrix}
    130.42 & 0 & 0\\ 0& 130.42 & 0\\ 0&0&130.42
    \end{bmatrix},
    \ m = 1.17, \ \alpha_{h} = 4.78, \\
    K &= \begin{bmatrix}
    538.18 & 0 & 0\\ 0& 538.18& 0\\ 0&0&538.18
    \end{bmatrix},  \  \beta_h = 17.67, \ \gamma_h = -68.95 
\end{aligned}
\end{equation}
where $m$, $K$, and $D$ units are in $kg$, $N/m$, and $N.s/m$ respectively, and rest of the parameters are constants.  

\vspace{-3 mm}
\subsection{Kinematic Validation}
\begin{figure}[t]
     \centering
     \begin{subfigure}[t]{0.15\textwidth}
     \caption{X tracking}
     \includegraphics[width=\textwidth]{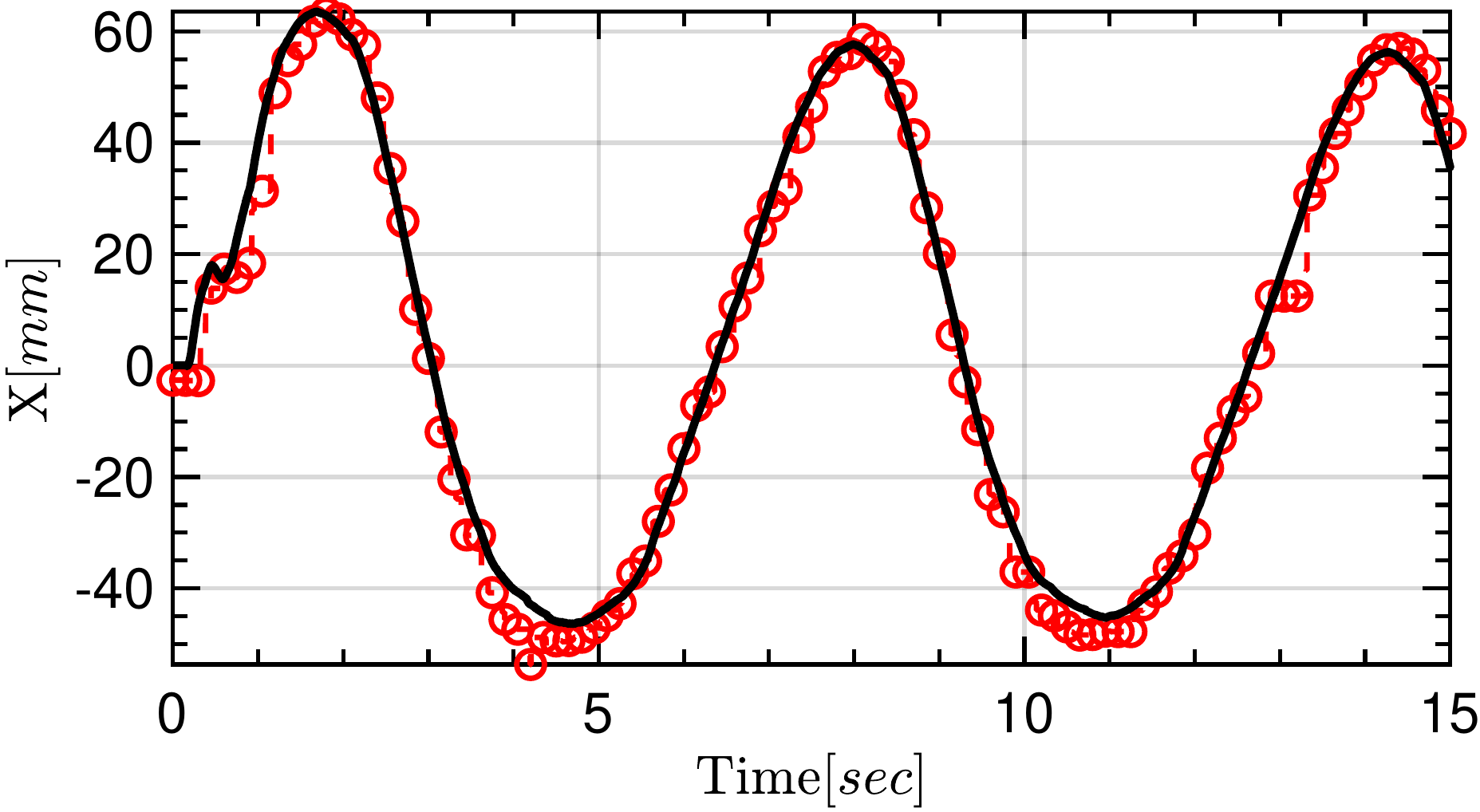}
     \end{subfigure}
     \begin{subfigure}[t]{0.15\textwidth}
     \caption{Y tracking}
     \includegraphics[width=\textwidth]{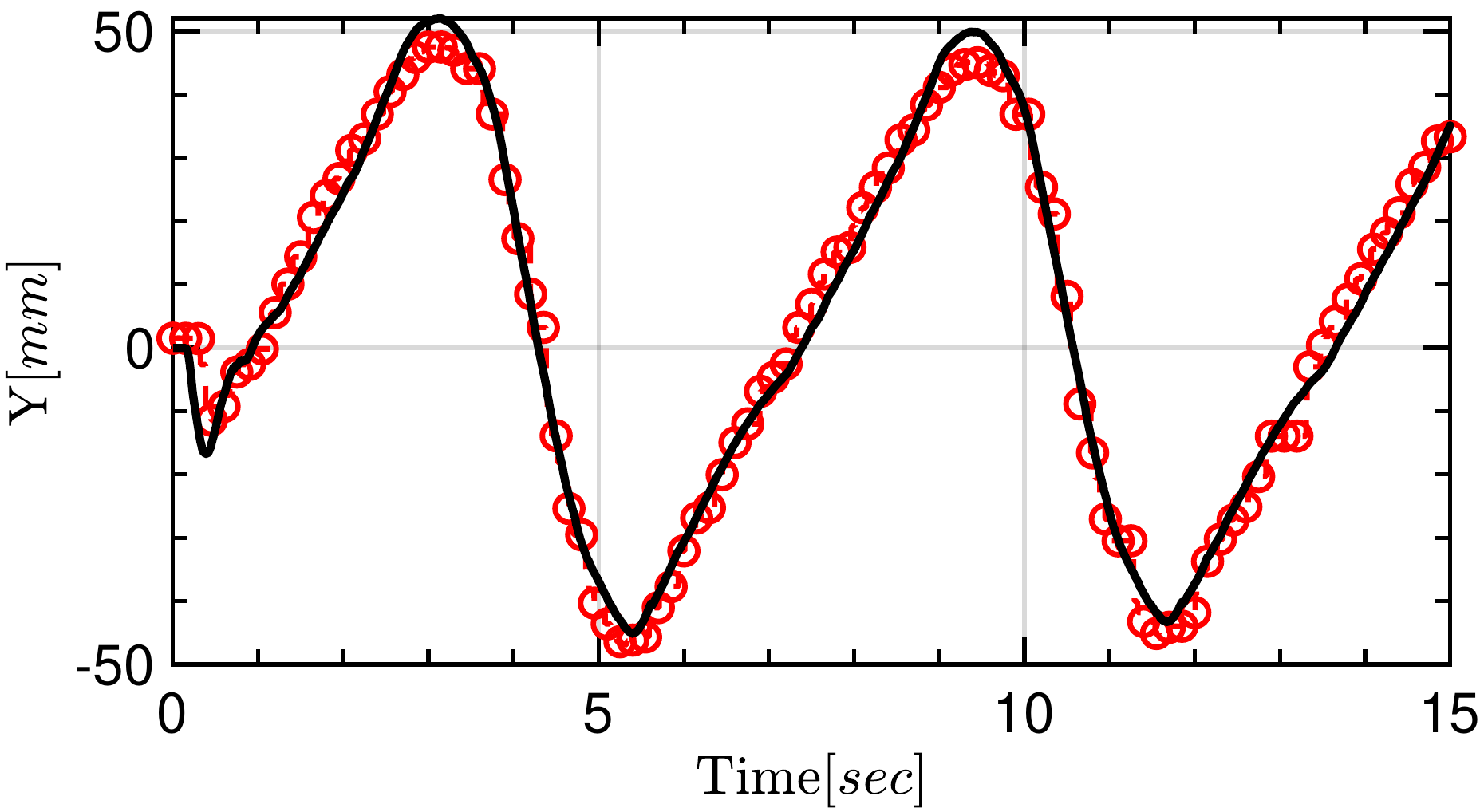}
     \end{subfigure}
     \begin{subfigure}[t]{0.15\textwidth}
      \caption{Z tracking}
     \includegraphics[width=\textwidth]{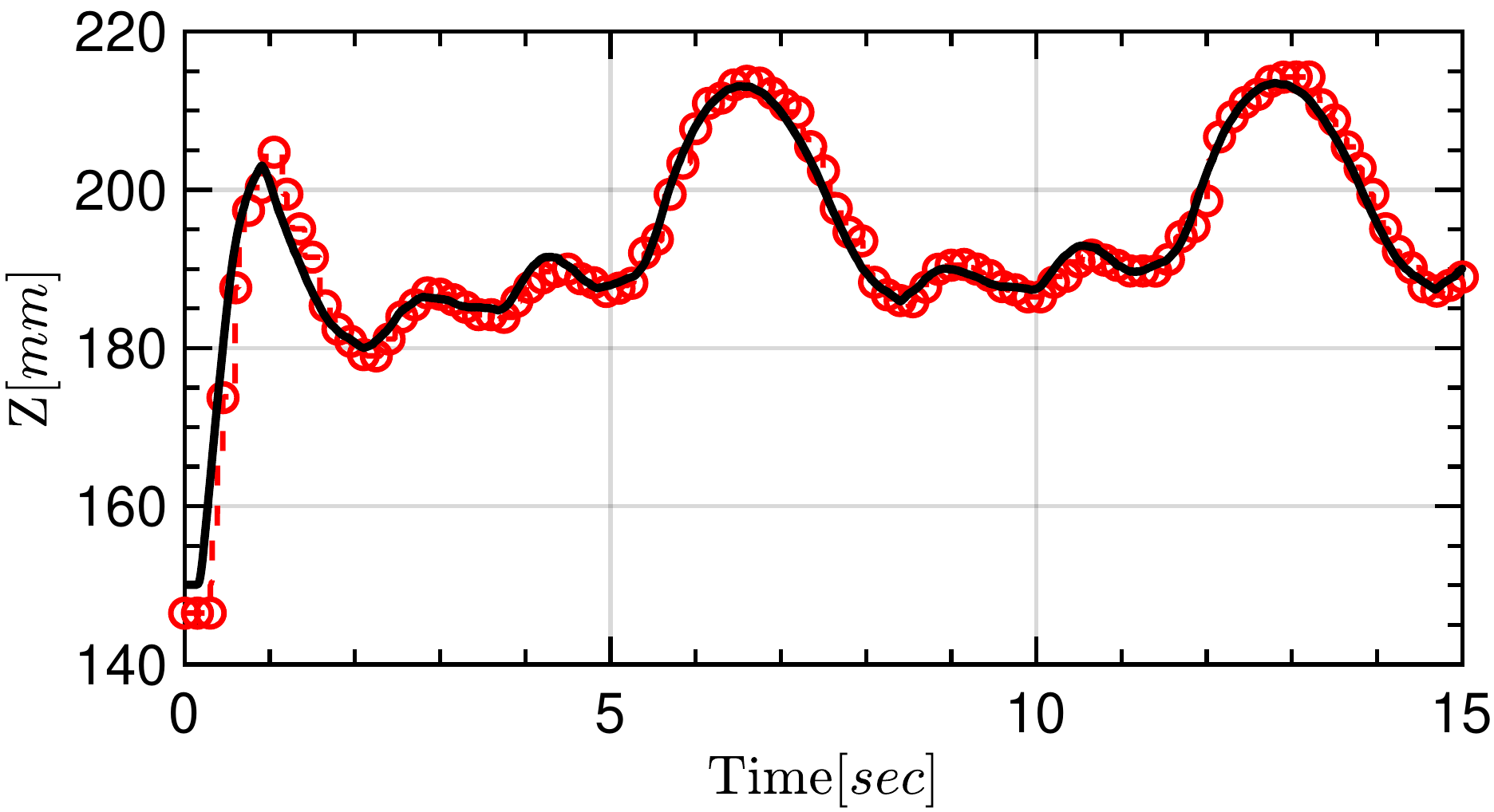}
     \end{subfigure}
     \begin{subfigure}[t]{0.24\textwidth}
      \caption{Position in XYZ}
     \includegraphics[width=\textwidth]{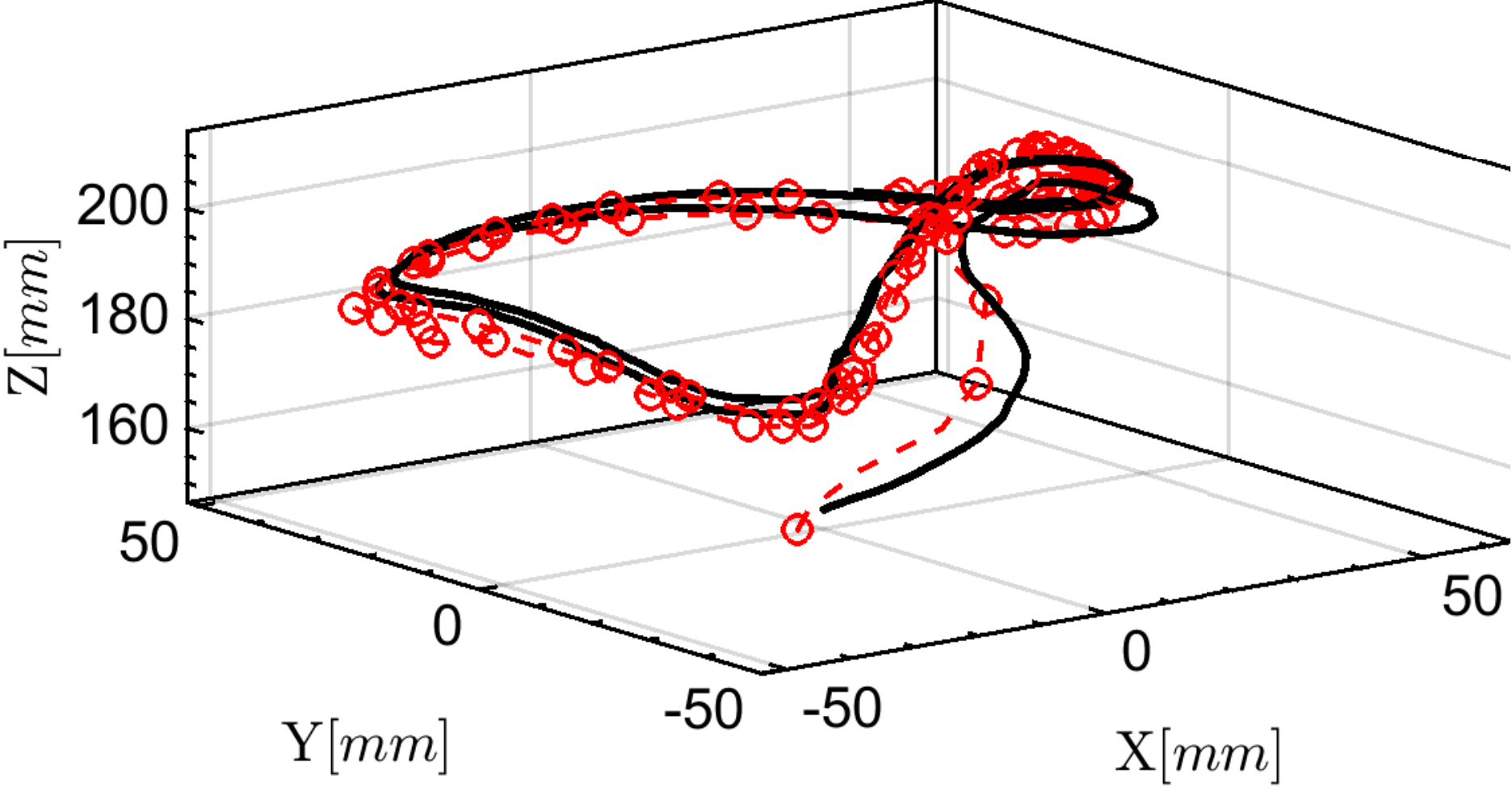}
     \end{subfigure}
     \begin{subfigure}[t]{0.24\textwidth}
      \caption{Task-space error}
     \includegraphics[width=\textwidth]{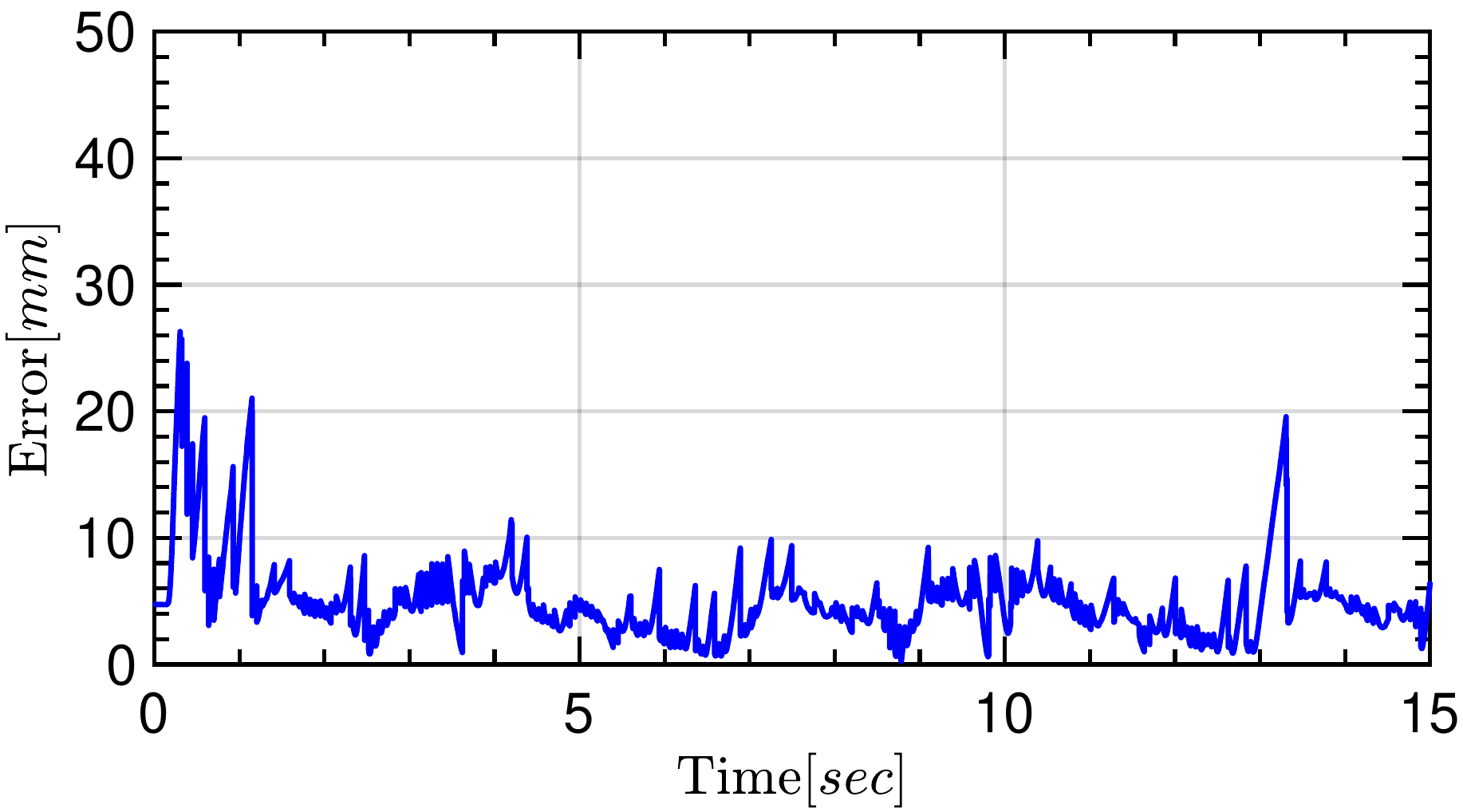}
     \end{subfigure}
     \caption{Kinematic model validation. (a-d) End-effector position in X-, Y-, Z-, and 3D space, respectively. Red indicates the measured results, and black indicates the modeled results. (e) The end-effector error plot. }
     \label{fig:PathOpenLoop1}
     \vspace{-7 mm}
\end{figure}

In this section, we would like to validate the kinematic mapping from actuator-space to task-space. To perform this experiment, we actuate each pneumatic muscle actuator based on the pre-defined signal input, calculate the task-space end-effector position, and measuring the end-effector via the MicronTracker camera, and compare the difference among the measured value and modeled value, as shown in Fig. \ref{fig:PathOpenLoop1}. The end-effector error is  5.06$\pm$3.25 mm, which is approximately 3.4\% of the robot length. The error is expected to come from 1) imperfect actuator length measurement, and 2) system dynamics, but the overall error is within the acceptable range. 

\vspace{-4 mm}
\subsection{Dynamic Trajectory Tracking}
In this section, the dynamic trajectory tracking performance of soft robotic arm will be investigated. The robot is required to follow a circular path, and the desired actuator variables are defined using closed-form inverse kinematic. %The closed-form solution is faster than the resolved-rate algorithm and it will eliminate any delay in path generation. To get the desired actuator variables
To achieve this, first, configuration variables will be calculated based on the desired position \cite{neppalli2009closed}. 
\begin{equation}
    \begin{aligned}
    \theta &= atan2(P_y, P_x), \
    \lambda = \frac{{P_x}^2+{P_y}^2+{P_z}^2}{2\sqrt{{P_x}^2+{P_y}^2}}, \
    \phi = asin(\frac{P_z}{\lambda})
    \end{aligned}
    \label{eq:Pos2Config}
\end{equation}
Using \eqref{eq:Pos2Config} and geometrical location of the actuators, the actuator-space variables can be calculated as follows \cite{godage2015modal} 
\begin{equation}
    l_{i} = (\lambda - r cos(\frac{2\pi}{3}(i-1) - \theta))\phi - L_0\\
    \label{eq:Config2Act}
\end{equation}
where $i \in [1,2,3]$.
To further highlight the performance of the system, the system behavior will be examined with various trajectory speed (1 $rad/s$, 3 $rad/s$), payload (200 g and 500 g,  and external disturbances (the payload is interfering with the end-effector). 

The system is required to follow a circle with a 5 cm radius in the XY plane as follows
\begin{equation}
    X = 0.05 sin(\omega t), \quad Y = 0.05 cos(\omega t), \quad Z = 0.19
\end{equation}
The control parameters are shown in Table \ref{tab:tab1}.

\begin{table}[htb!]
%    \rowcolors{1}{gray}{lightgray}
    \centering
        \begin{tabular}{|c|c|c|c|c|c|c|}
        \hline
             Parameter & $\Lambda_{AP}$ & $k_g$ & $k_p$ & $k_d$ &${\Gamma_K}$ &$\Gamma_D$\\
        \hline
            Value & 8 & 8 & $30^2$ & 60 & $1e^{-5}$ & $1e^{-2}$ \\
        \hline
        \end{tabular}
    \caption{The control parameters for AP and PDFL}
    \label{tab:tab1}
\end{table}

\subsubsection{Trajectory tracking with the speed of $1 \ rad/s$ and no load}
In the first scenario, the system will follow a path with the speed of $1 \ rad/s $, where the high-frequency dynamic terms don't have a high impact on the system. The results in both actuator-space  and task-space are provided to  highlight the system performance. Note that the task space results are based on encoder measurements and forward kinematics since the camera is not fast enough and will result in interrupted data from marker readings. It should be noted that incorporating a robust term such as switching term \cite{slotine1987adaptive}, or increasing the control gains could result in better accuracy but the chattering will generate non-smooth behavior and oscillation in the robot.

\begin{figure}[t]
     \centering
     \begin{subfigure}[c]{0.15\textwidth}
      \caption{$l_1$ tracking}
     \includegraphics[width=\textwidth]{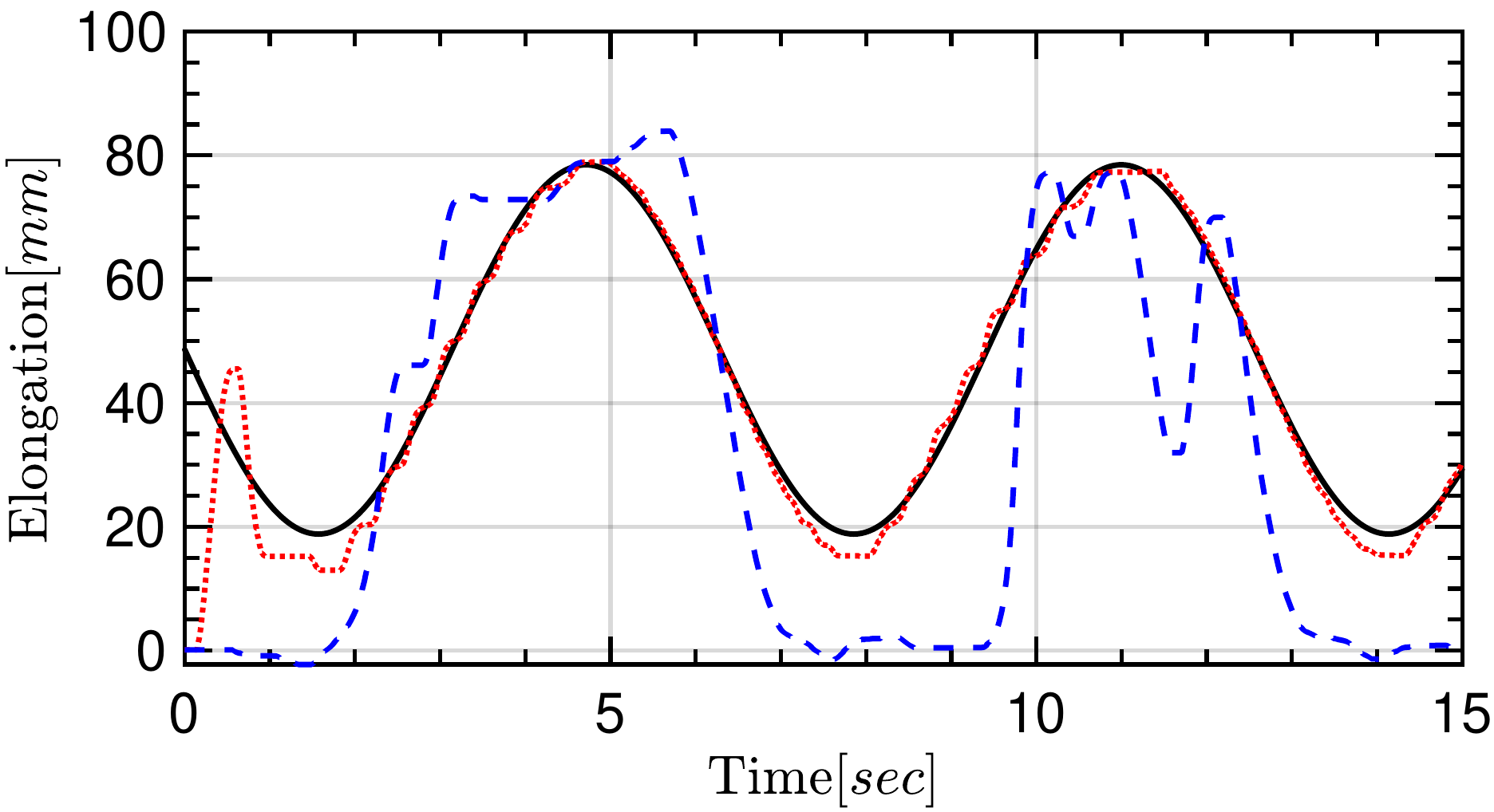}
     \end{subfigure}
     \begin{subfigure}[c]{0.15\textwidth}
      \caption{$l_2$ Tracking}
     \includegraphics[width=\textwidth]{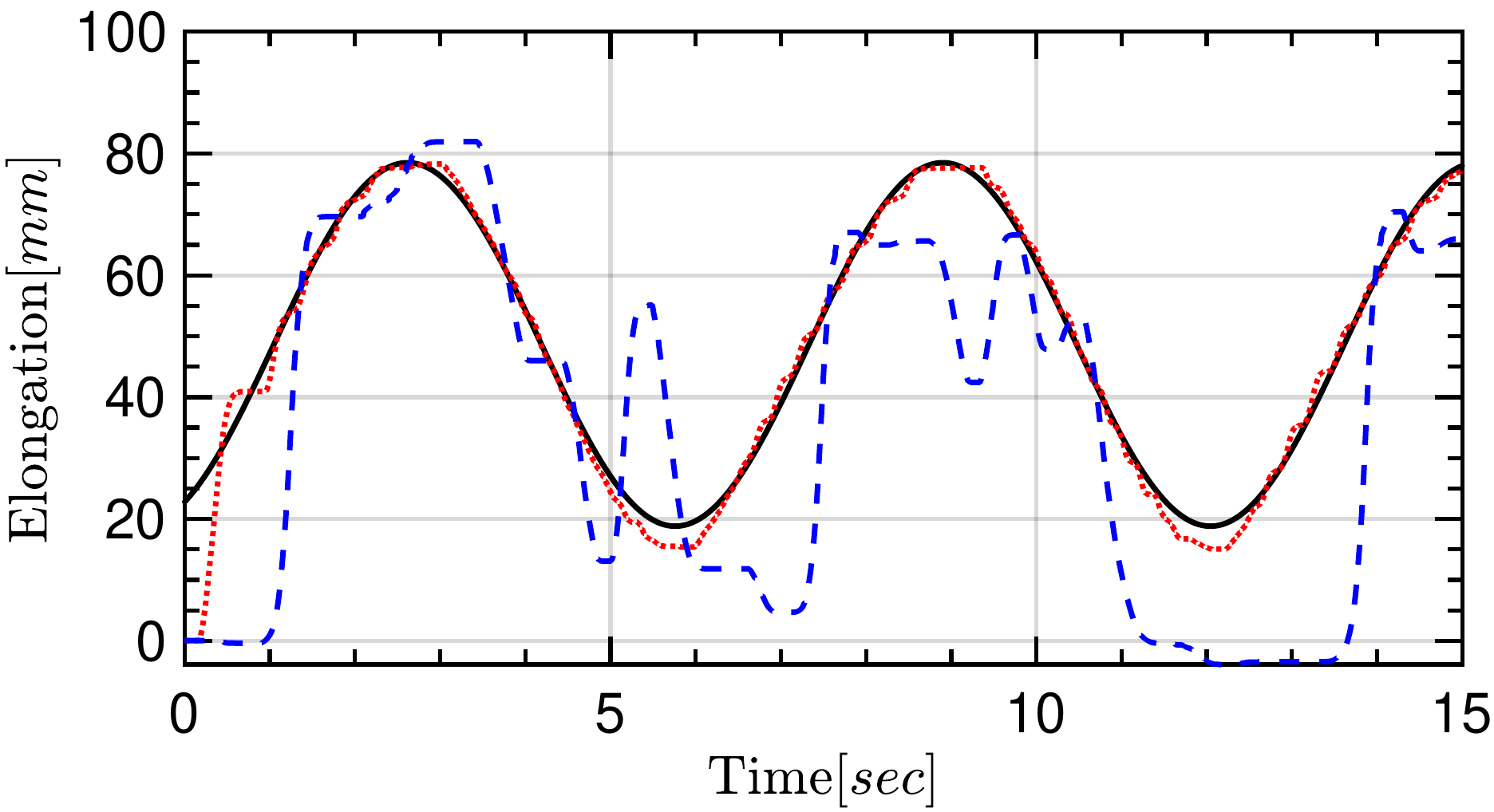}
     \end{subfigure}
     \begin{subfigure}[c]{0.15\textwidth}
      \caption{$l_3$ Tracking}
     \includegraphics[width=\textwidth]{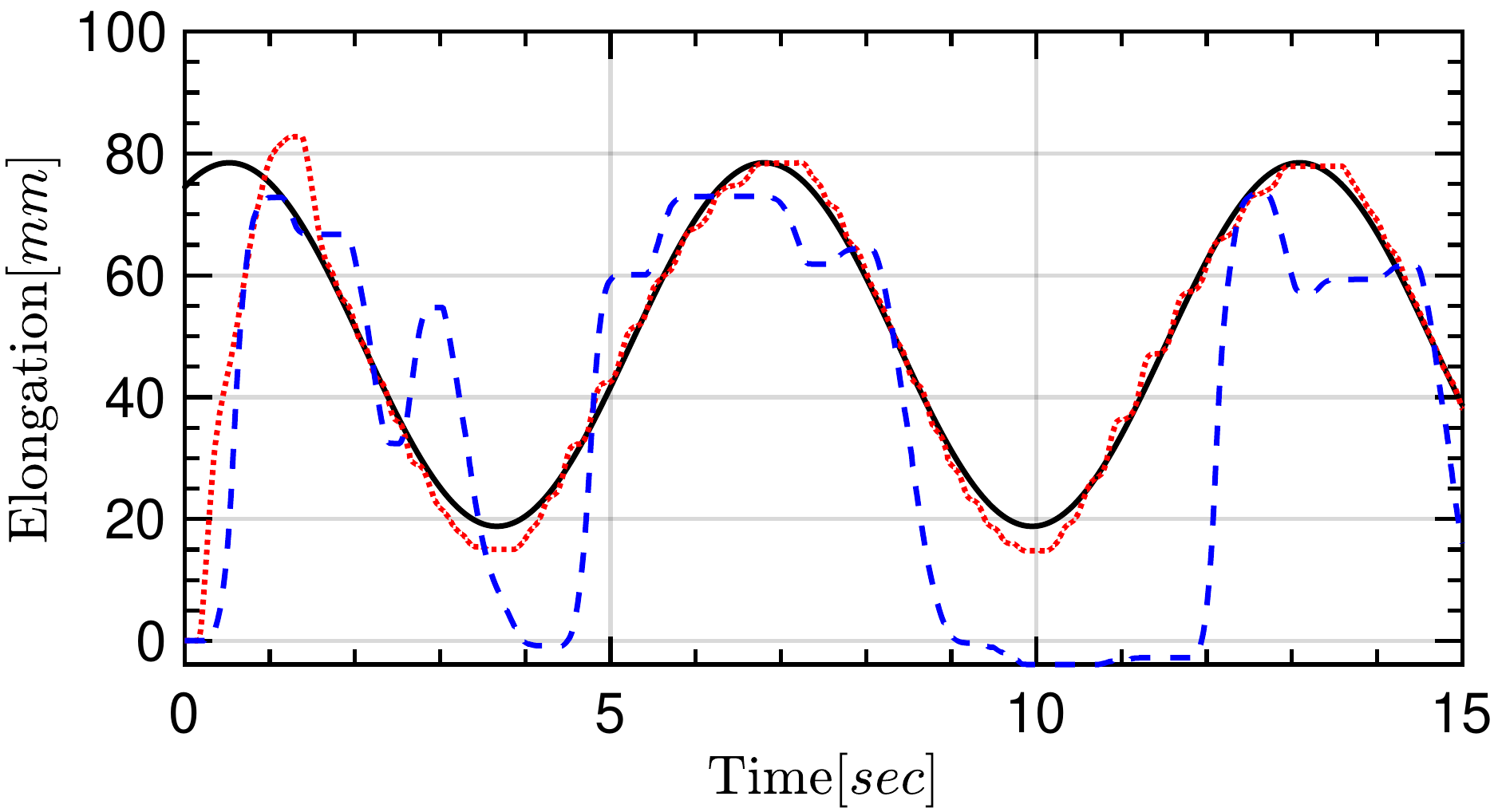}
     \end{subfigure}
     \begin{subfigure}[c]{0.24\textwidth}
      \caption{Position tracking in XY plane}
     \includegraphics[width=\textwidth]{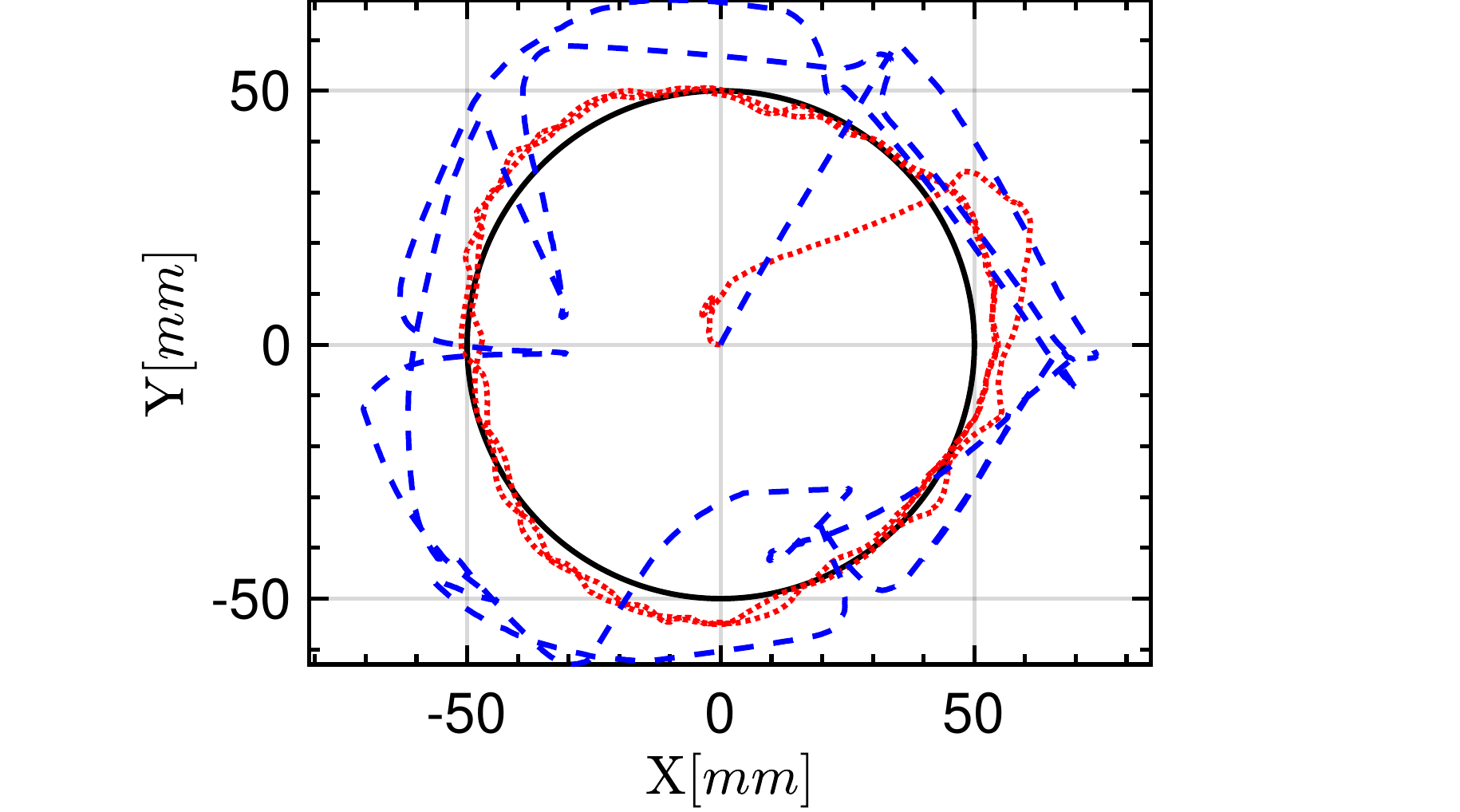}
     \end{subfigure}
     \begin{subfigure}[c]{0.24\textwidth}
      \caption{Position tracking in XYZ plane}
     \includegraphics[width=\textwidth]{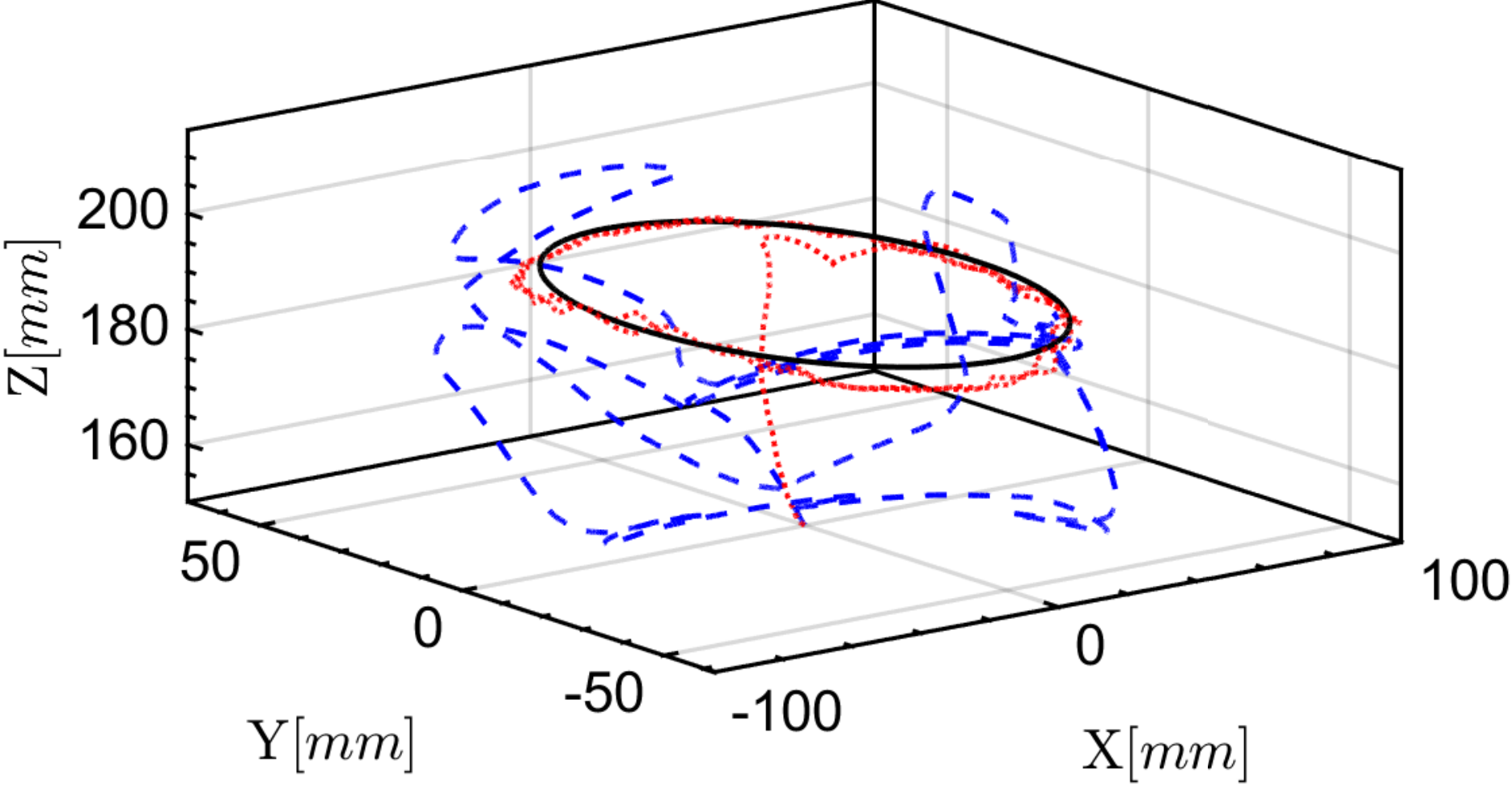}

     \end{subfigure}
     \caption{Trajectory tracking in actuator space and task space. The AP is in red, PDFL in blue and the desired trajectory in black with $\omega = 1 \ rad/s$.}
     \label{fig:PathTrack1radfree}
\end{figure}

As can be seen in Fig. \ref{fig:PathTrack1radfree}, the AP control presented a very smooth response with good accuracy. After the transient response, where the adaptive parameters are updated, the steady-state response could accurately follow the desired trajectory, while the PDFL presents both overshoot and undershoot, and lost performance at different states. 
\begin{figure}[t]
     \centering
     \begin{subfigure}[c]{0.15\textwidth}
     \caption{$l_1$ Tracking}
     \includegraphics[width=\textwidth]{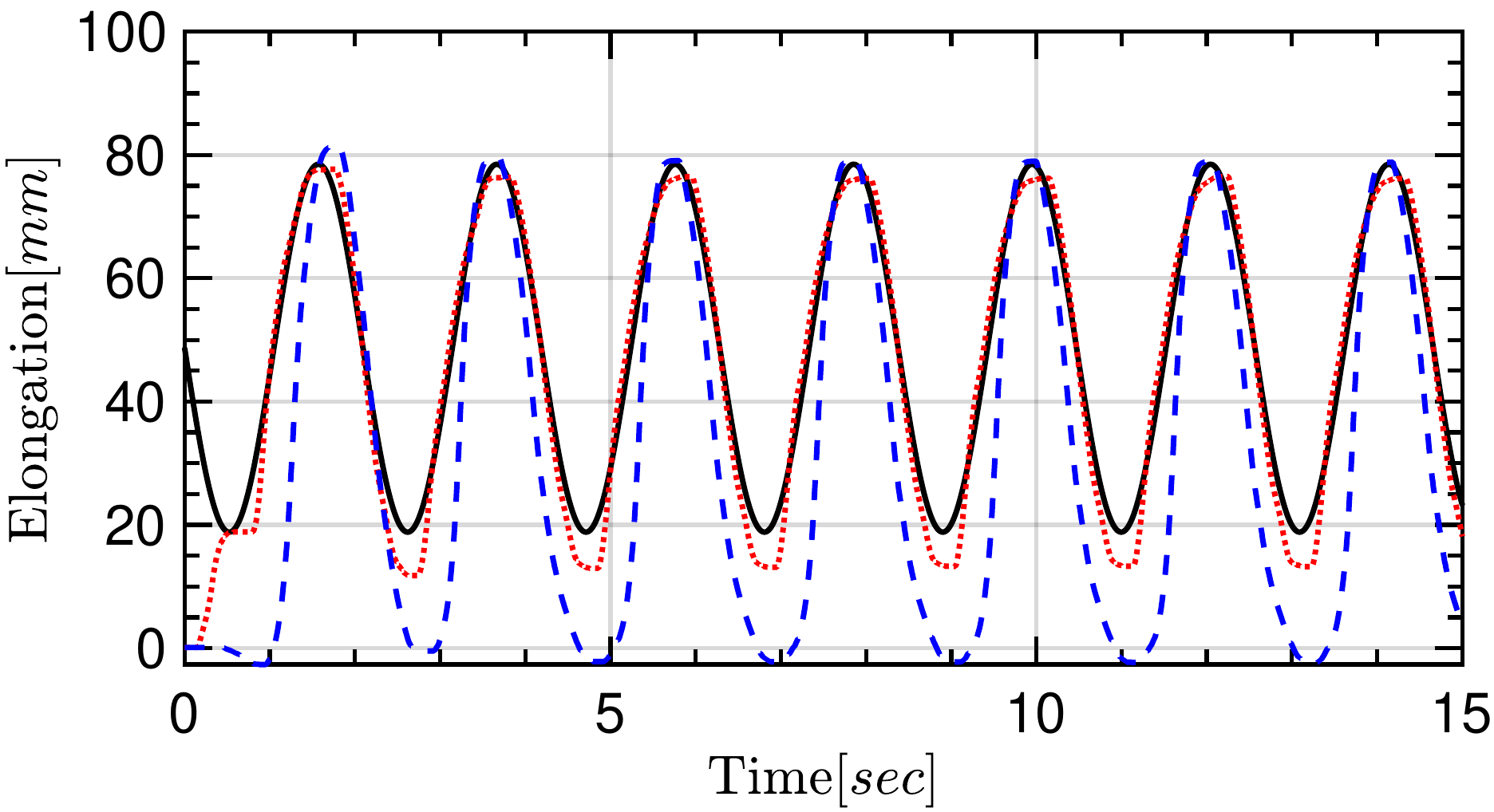}
     \end{subfigure}
     \begin{subfigure}[c]{0.15\textwidth}
      \caption{$l_2$ Tracking}
     \includegraphics[width=\textwidth]{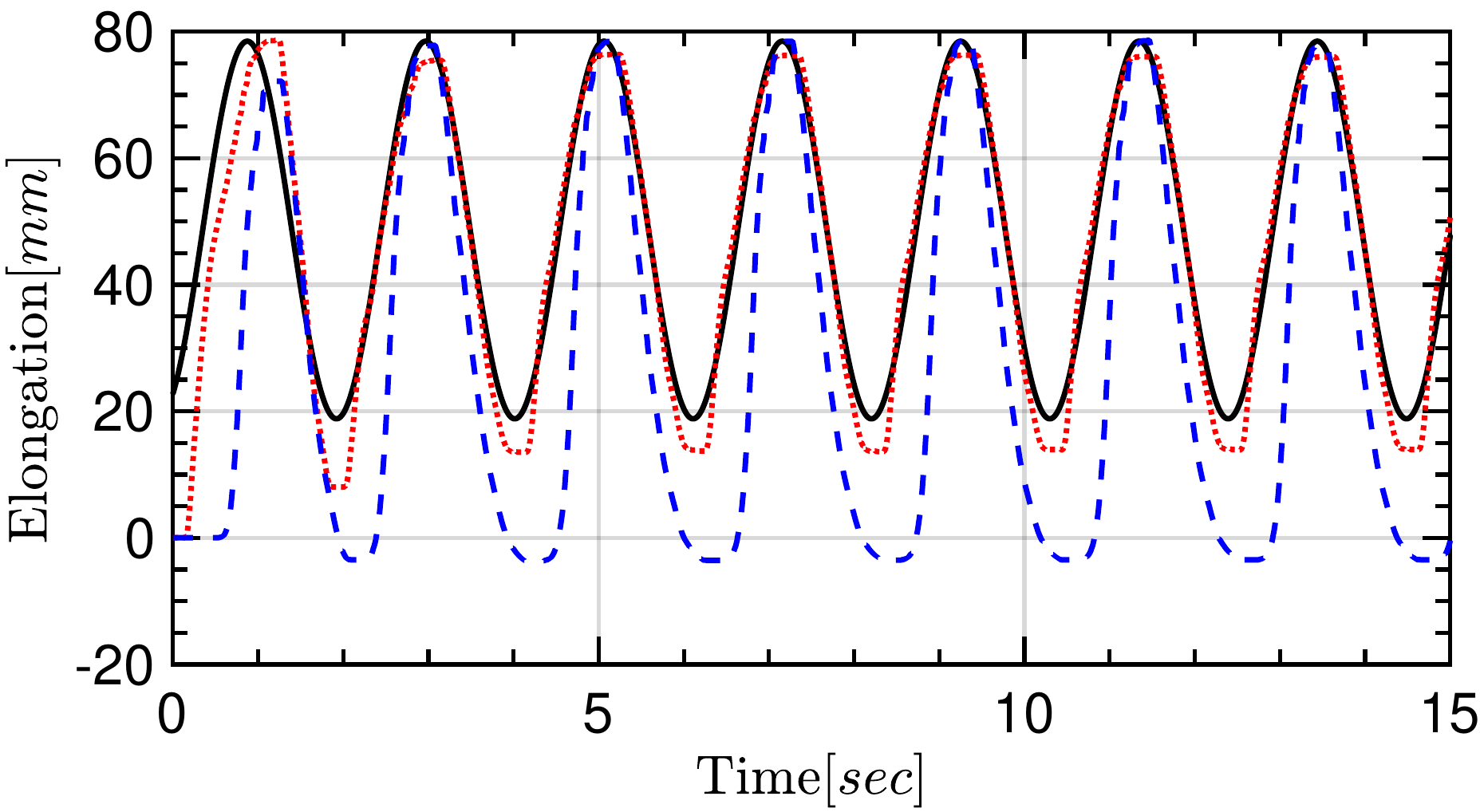}
     \end{subfigure}
     \begin{subfigure}[c]{0.15\textwidth}
      \caption{$l_3$ Tracking}
     \includegraphics[width=\textwidth]{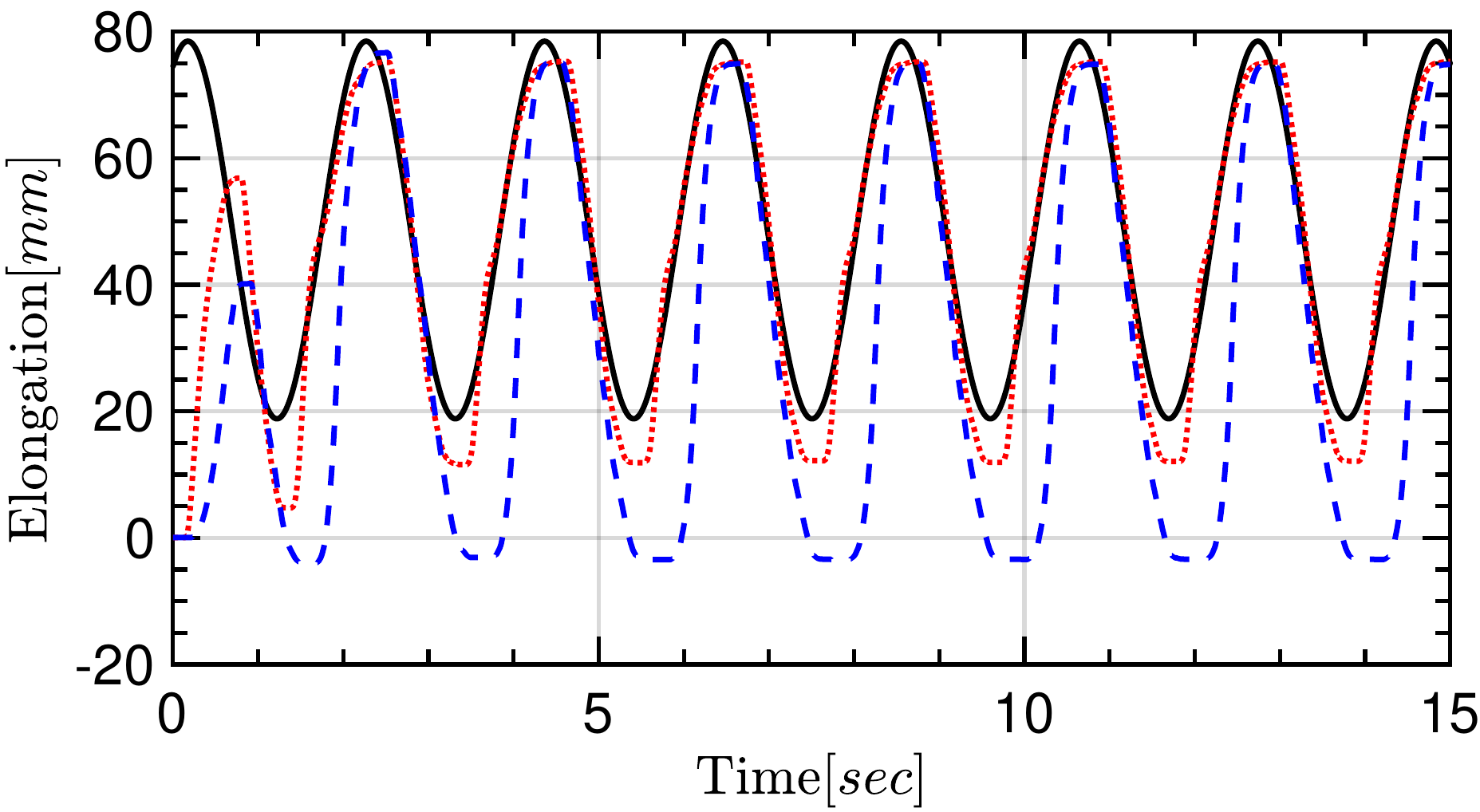}
     \end{subfigure}
     \begin{subfigure}[c]{0.24\textwidth}
      \caption{Position tracking in XY plane}
     \includegraphics[width=\textwidth]{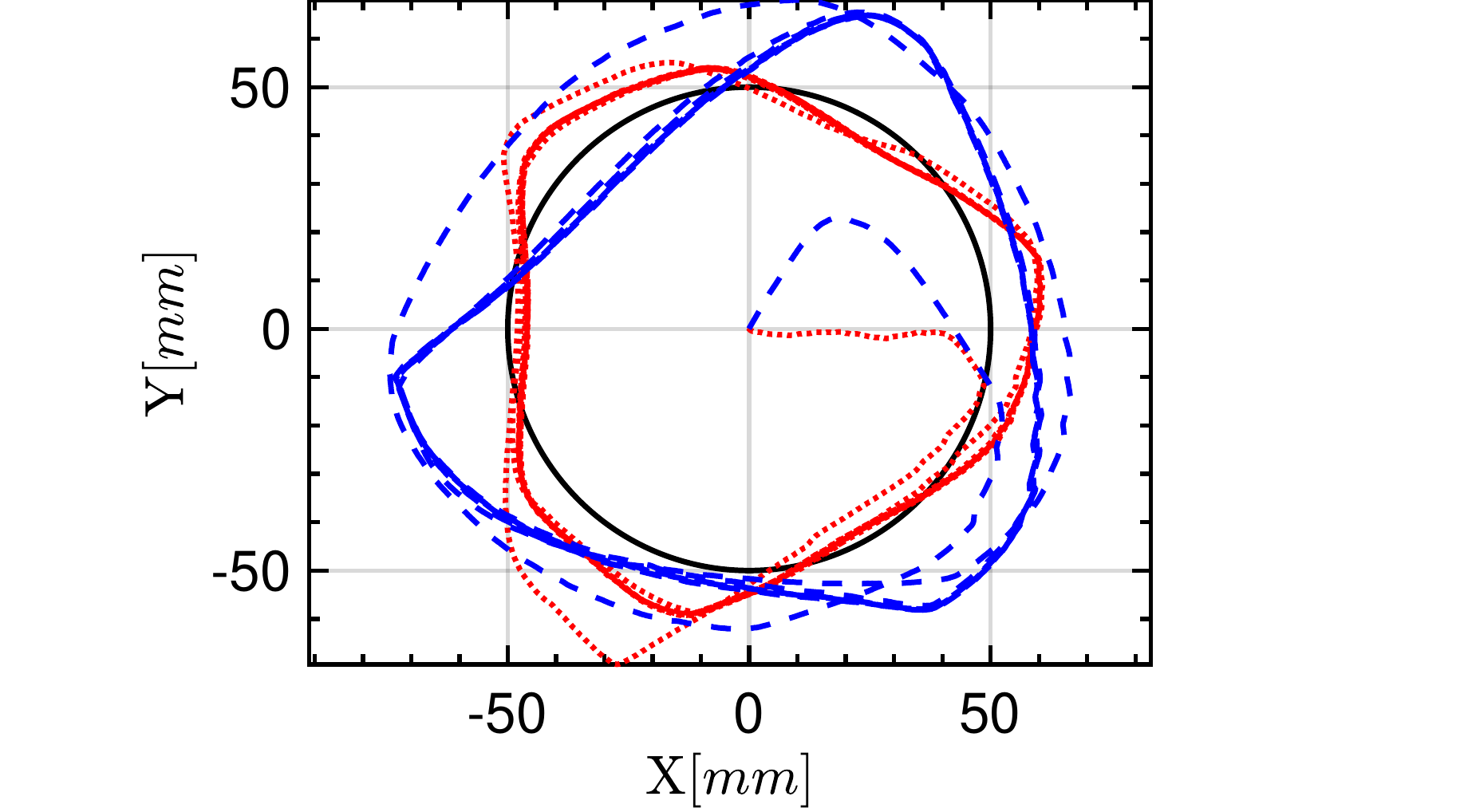}
     \end{subfigure}
     \begin{subfigure}[c]{0.24\textwidth}
      \caption{Position tracking in XYZ plane}
     \includegraphics[width=\textwidth]{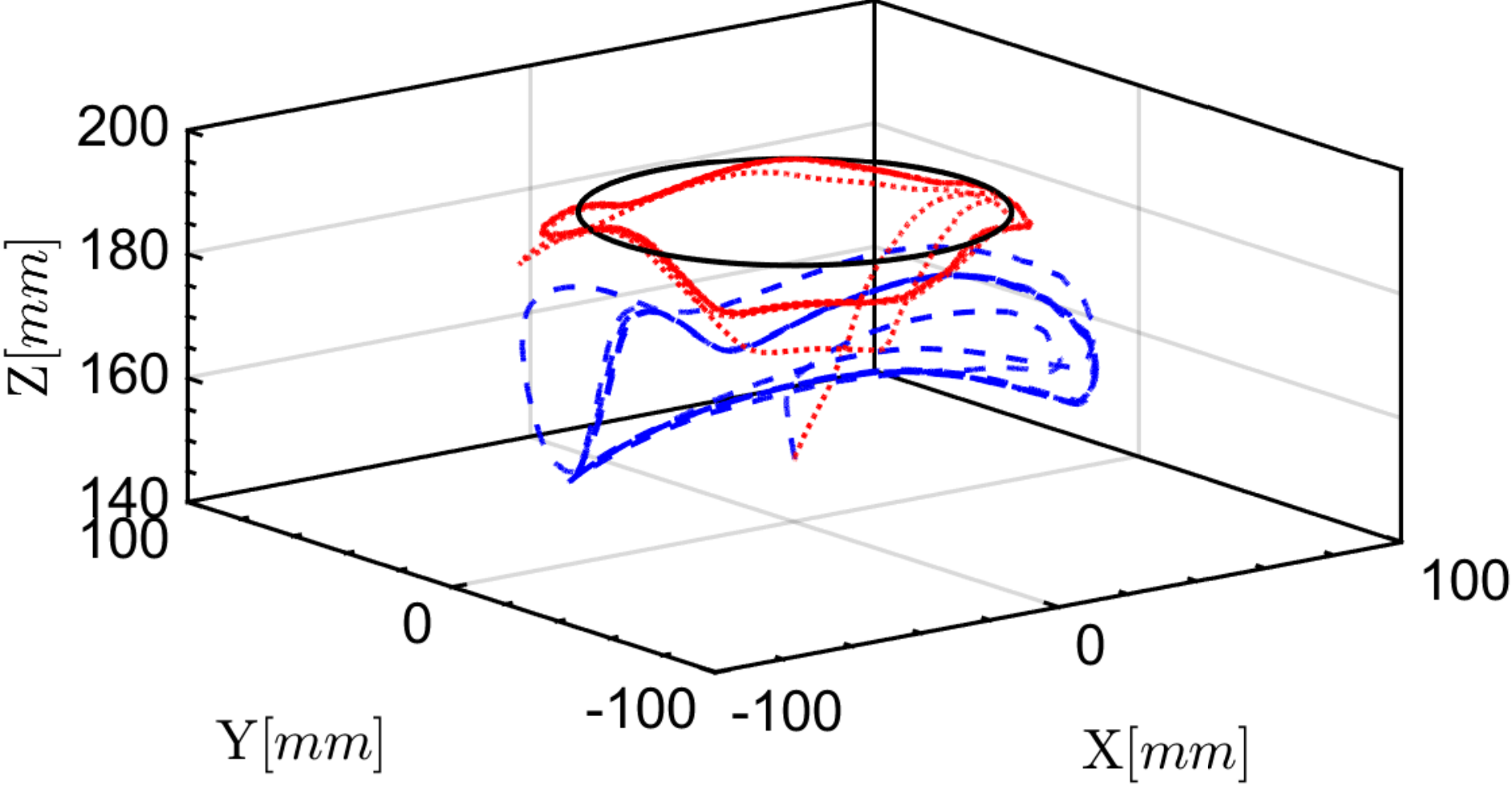}
     \end{subfigure}
     \caption{Trajectory tracking in actuator space and task space. The AP is in red, PDFL in blue and the desired trajectory in black with $\omega = 3 \ rad/s$.}
     \label{fig:PathTrack3radfree}
\end{figure}
\subsubsection{Trajectory tracking   with the speed of $3 \ rad/s$ and no load}
\label{sub:3radfree}
As depicted in Fig. \ref{fig:PathTrack3radfree}, the tracking error has also increased with the increment of the speed. However, we would like to point out that the system maintained its stability with AP control.

\subsubsection{Trajectory tracking with the speed of $1 \ rad/s$ subjected to a 200 g payload and external disturbances}

\begin{figure}[t]
     \centering
     \begin{subfigure}[c]{0.15\textwidth}
          \caption{$l_1$ Tracking}
     \includegraphics[width=\textwidth]{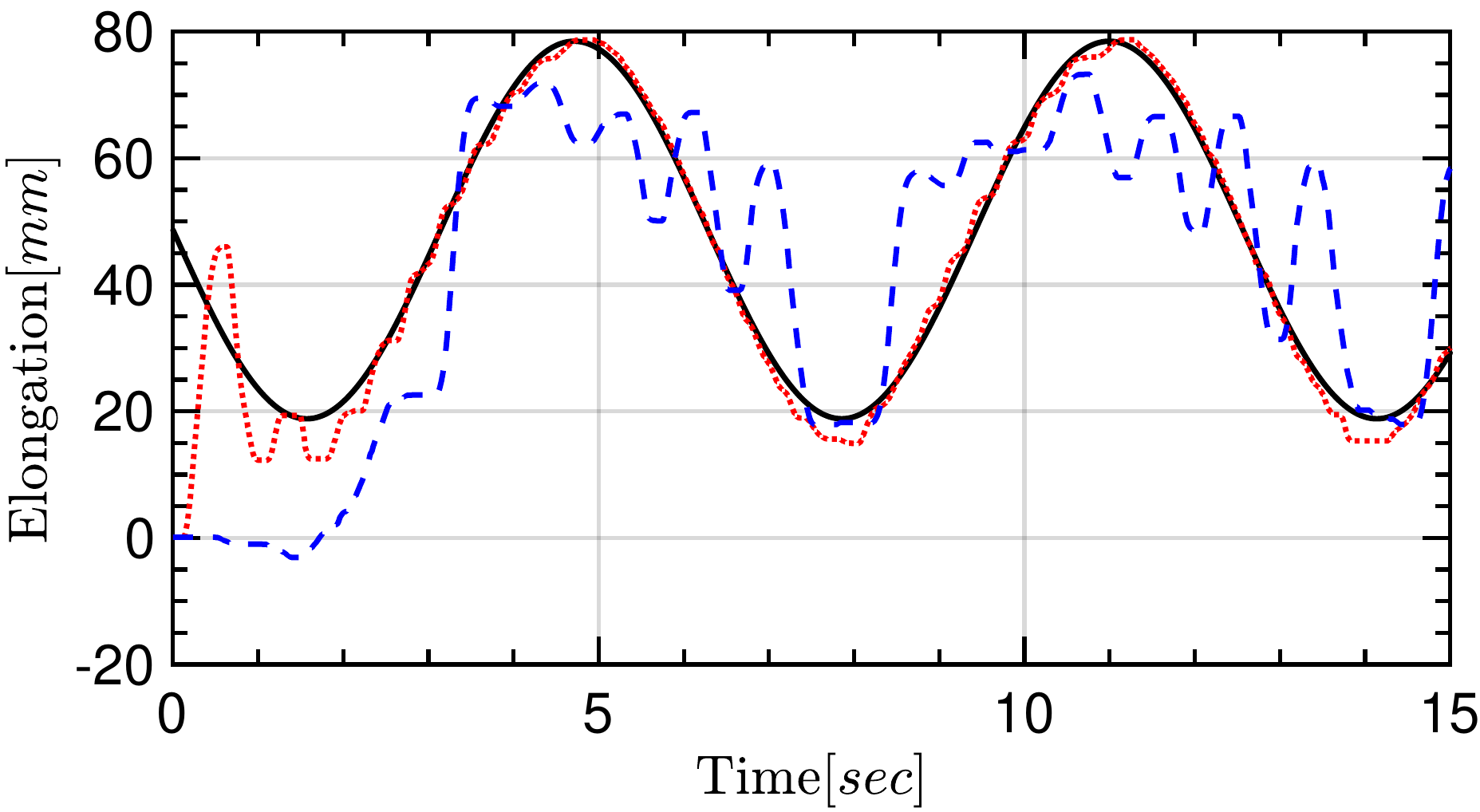}
     \end{subfigure}
     \begin{subfigure}[c]{0.15\textwidth}
          \caption{$l_2$ Tracking}
     \includegraphics[width=\textwidth]{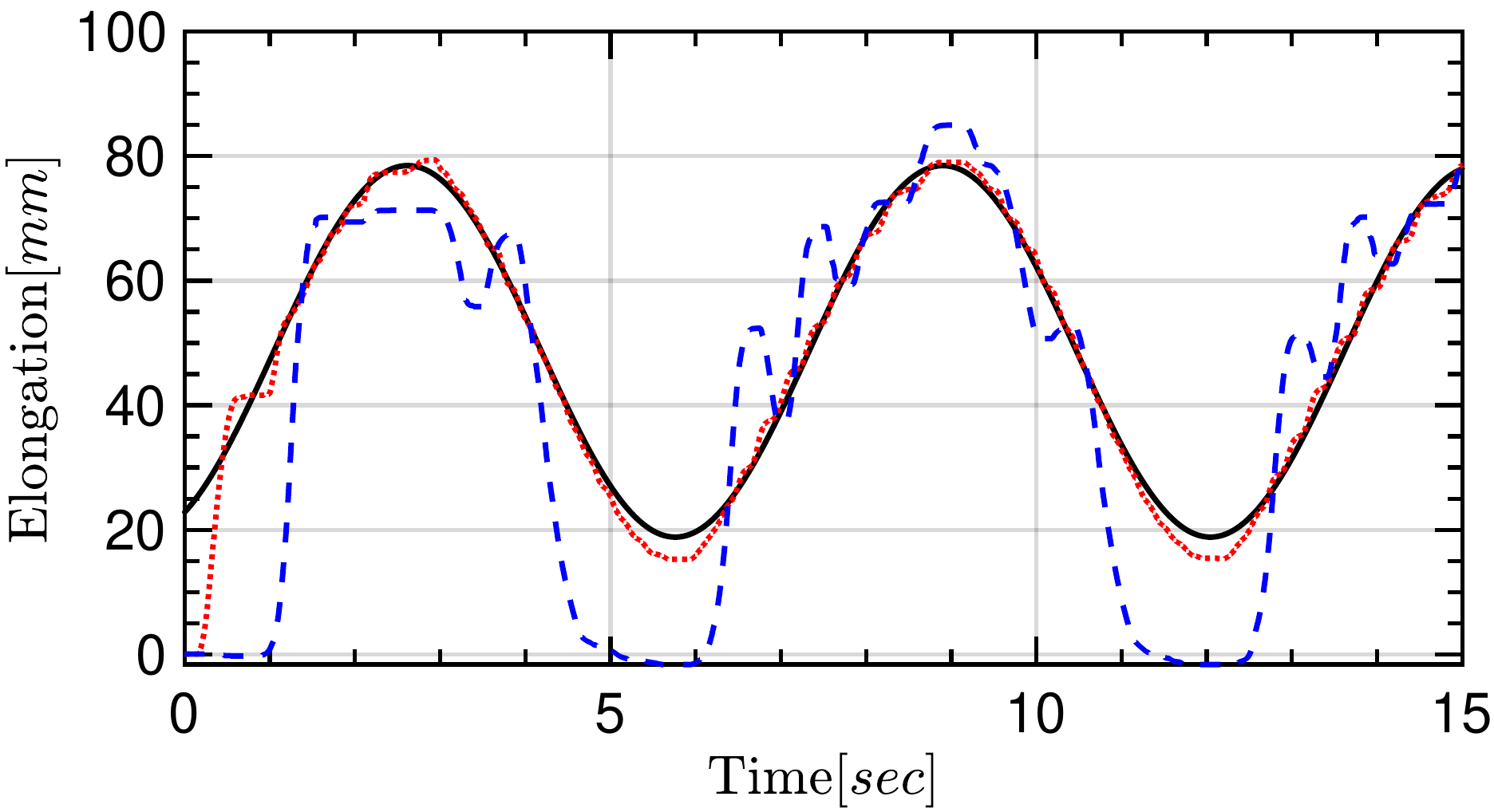}

     \end{subfigure}
     \begin{subfigure}[c]{0.15\textwidth}
          \caption{$l_3$ Tracking}
     \includegraphics[width=\textwidth]{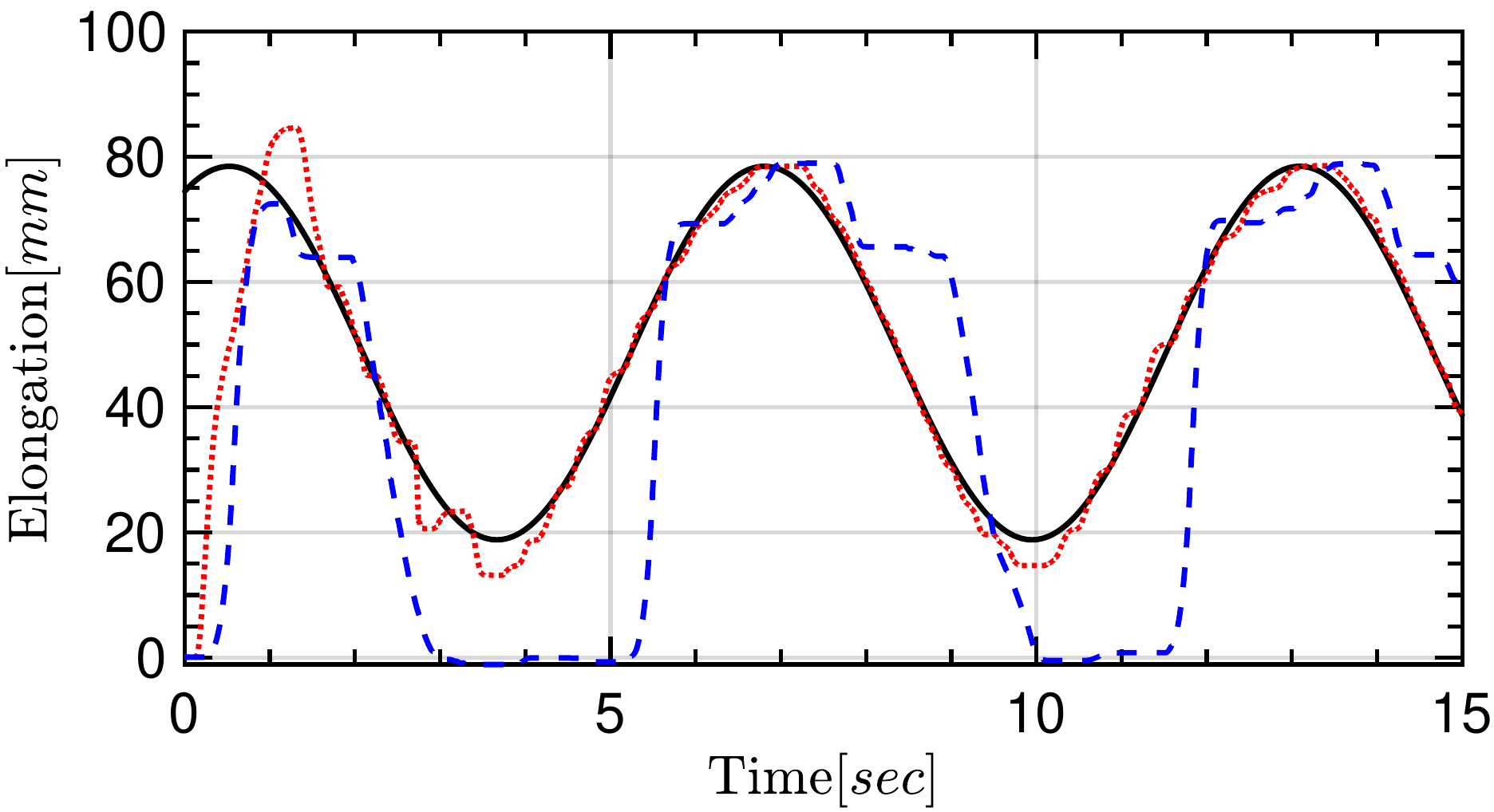}

     \end{subfigure}
     \begin{subfigure}[c]{0.24\textwidth}
          \caption{Position tracking in XY plane}
     \includegraphics[width=\textwidth]{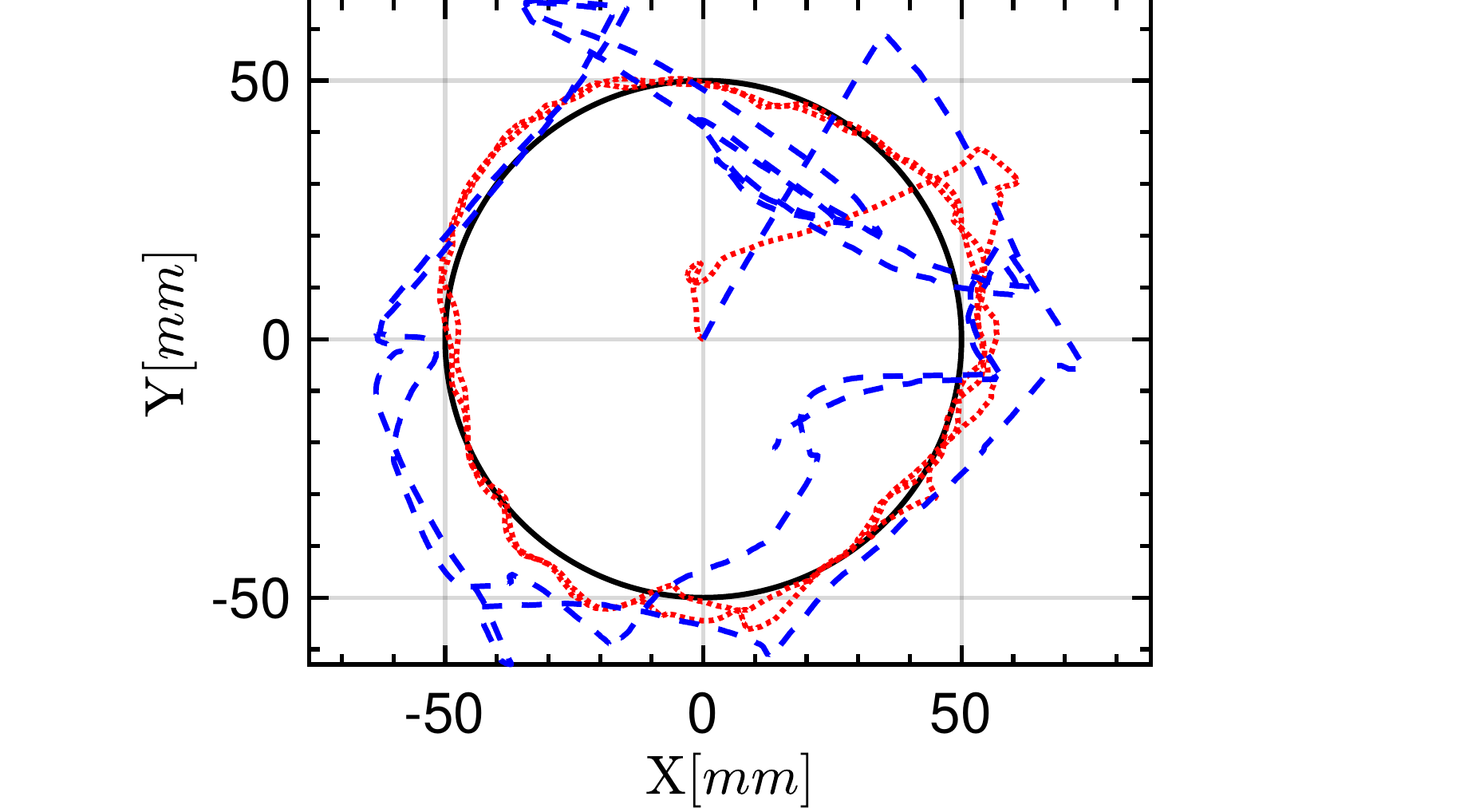}

     \end{subfigure}
     \begin{subfigure}[c]{0.24\textwidth}
          \caption{Position tracking in XYZ plane}
     \includegraphics[width=\textwidth]{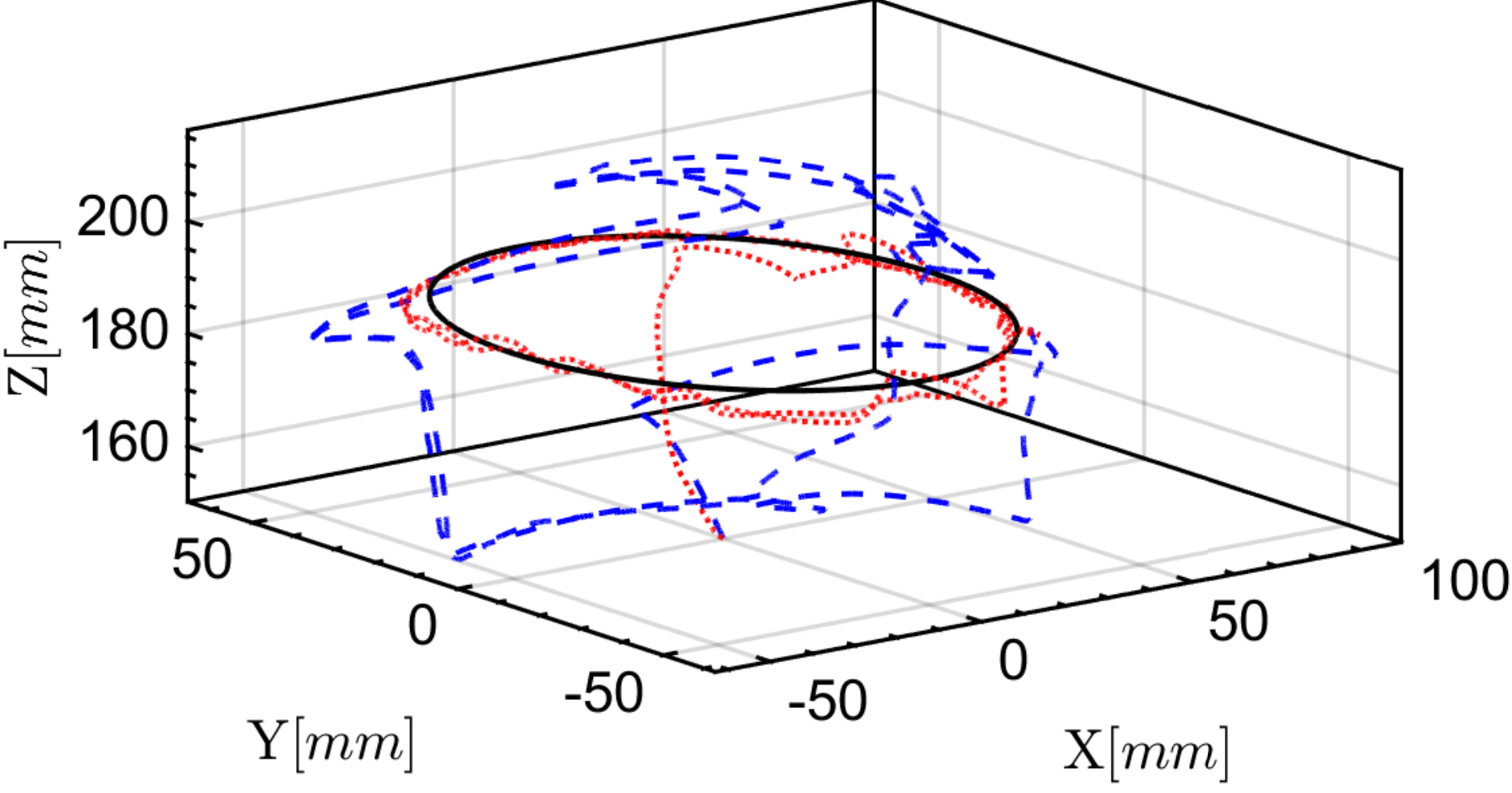}

     \end{subfigure}
     \caption{Trajectory tracking in actuator space and task space. The AP is in red, PDFL in blue and the desired trajectory in black with $\omega = 1 \ rad/s$. The system is subjected to a moving 200g load.}
     \label{fig:PathTrack1radLoad200g}
     \vspace{-6 mm}
\end{figure}

In this section, trajectory tracking was performed with the consideration of external load and  unknown disturbance. 
A 200 g payload is loosely attached to one of the holes on the end plate. The loose connection style was intentionally chosen such that the load will move in a random style during operation, pushing the limit of the proposed AP controller.
The results are depicted in Fig. \ref{fig:PathTrack1radLoad200g}. 
As can be seen, the AP controller maintained its tracking performance while the PDFL couldn't handle the scenario with  load/disturbance. This behavior highlights the disturbance rejection ability of the proposed AP control.

\subsubsection{Trajectory tracking with the speed of $3 \ rad/s$ subjected to a 200 g payload and external disturbances}

\begin{figure}[t]
     \centering
     \begin{subfigure}[c]{0.15\textwidth}
          \caption{$l_1$ Tracking}
     \includegraphics[width=\textwidth]{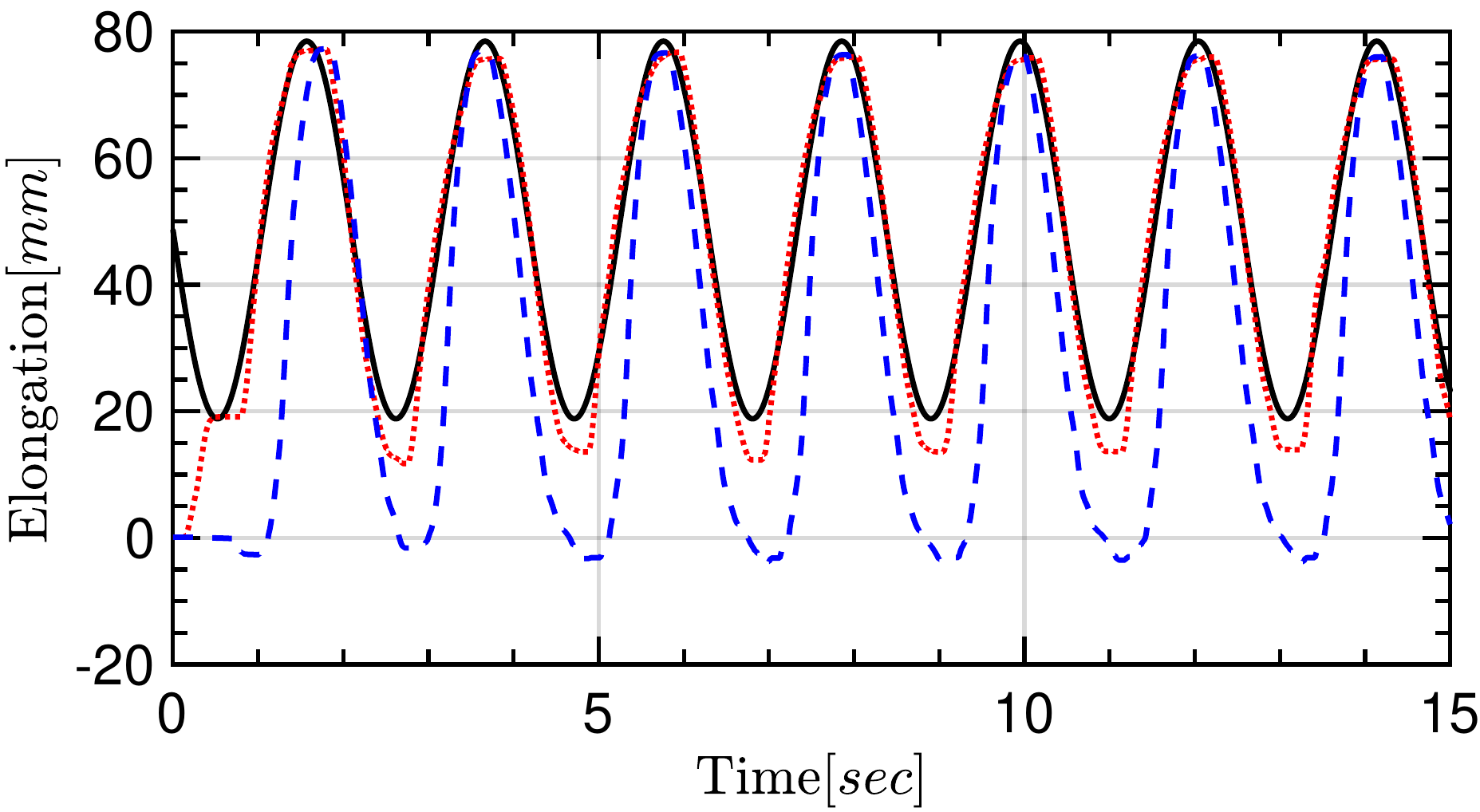}

     \end{subfigure}
     \begin{subfigure}[c]{0.15\textwidth}
          \caption{$l_2$ Tracking}
     \includegraphics[width=\textwidth]{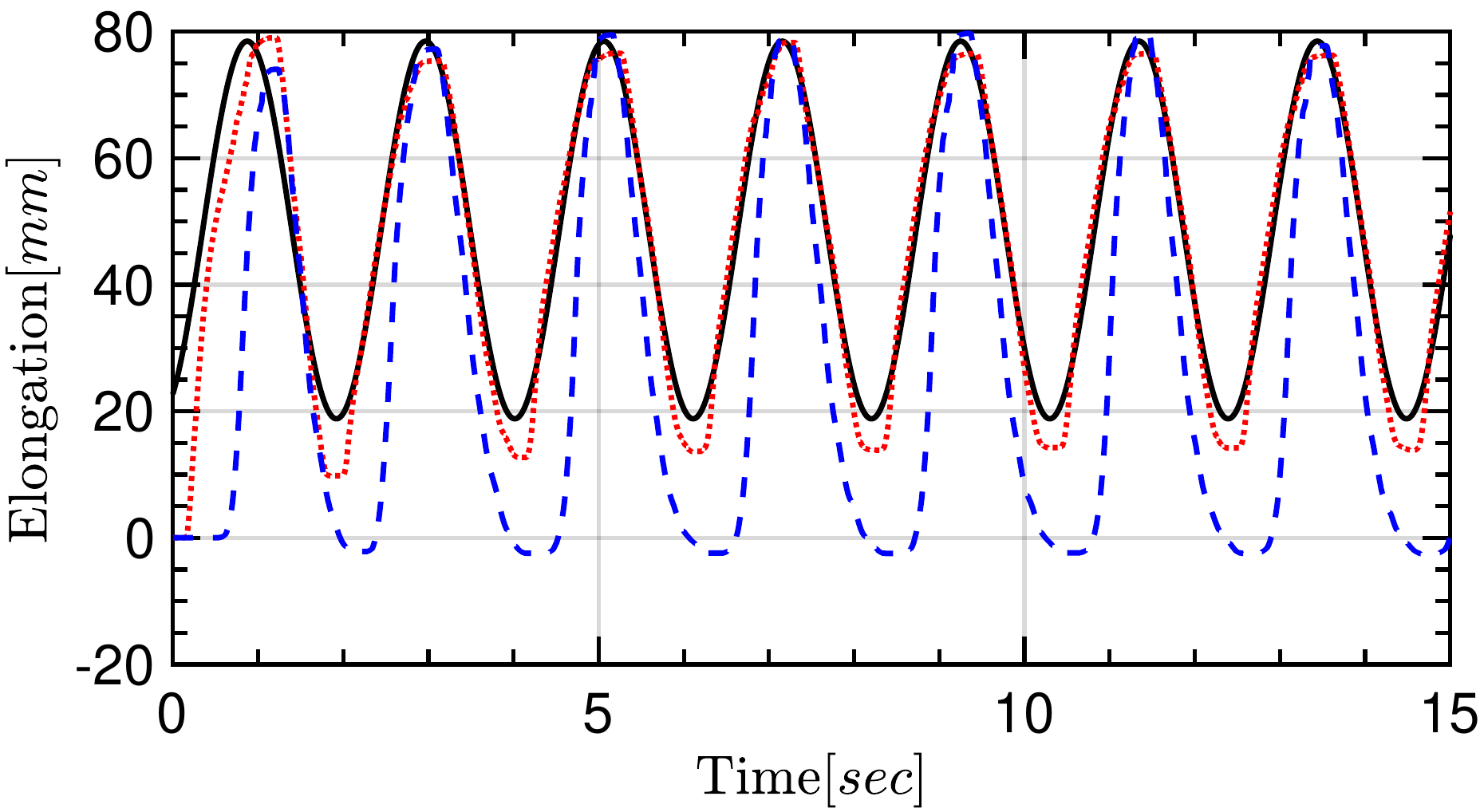}

     \end{subfigure}
     \begin{subfigure}[c]{0.15\textwidth}
          \caption{$l_3$ Tracking}
     \includegraphics[width=\textwidth]{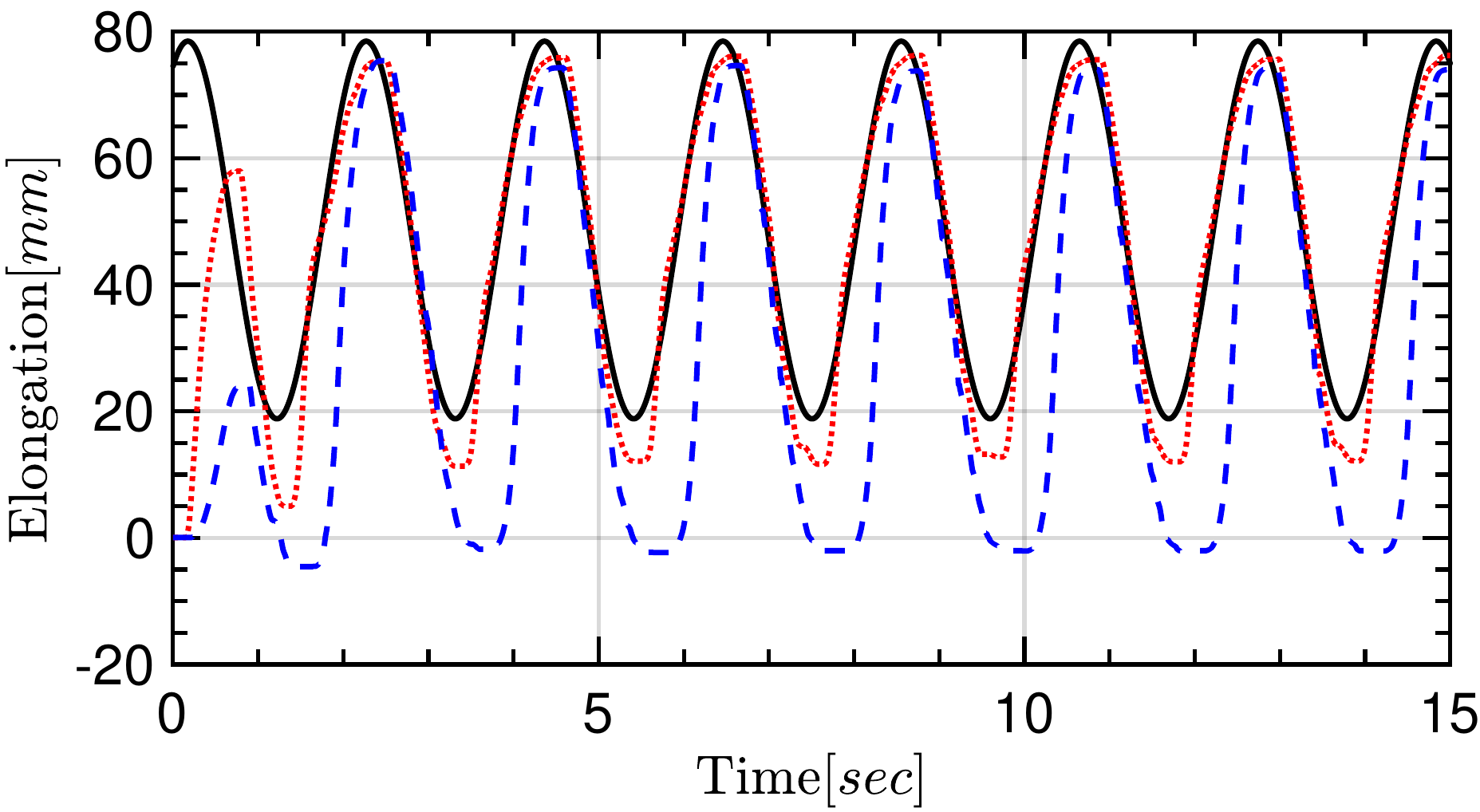}

     \end{subfigure}
     \begin{subfigure}[c]{0.24\textwidth}
          \caption{Position tracking in XY plane}
     \includegraphics[width=\textwidth]{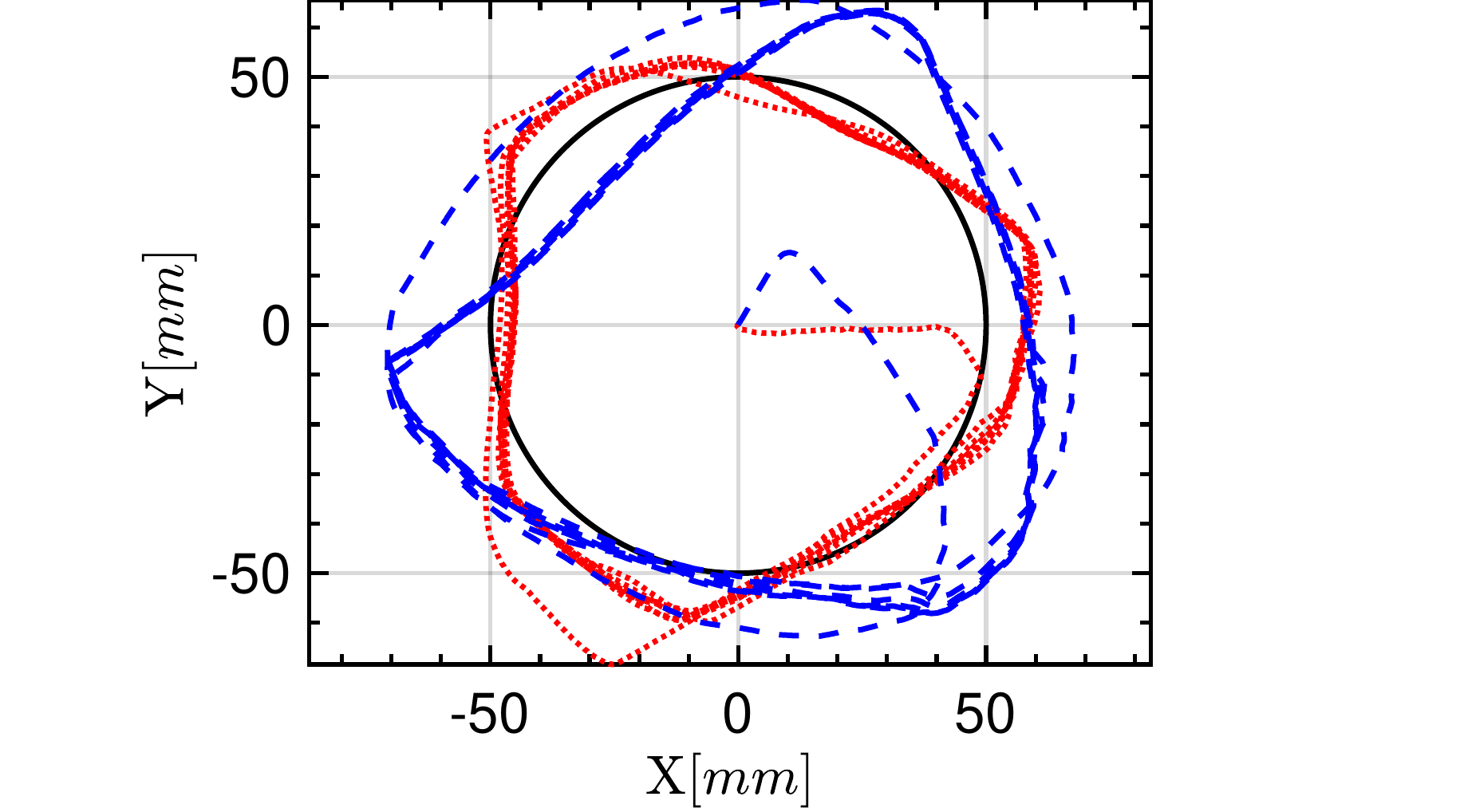}

     \end{subfigure}
     \begin{subfigure}[c]{0.24\textwidth}
          \caption{Position tracking in XYZ plane}
     \includegraphics[width=\textwidth]{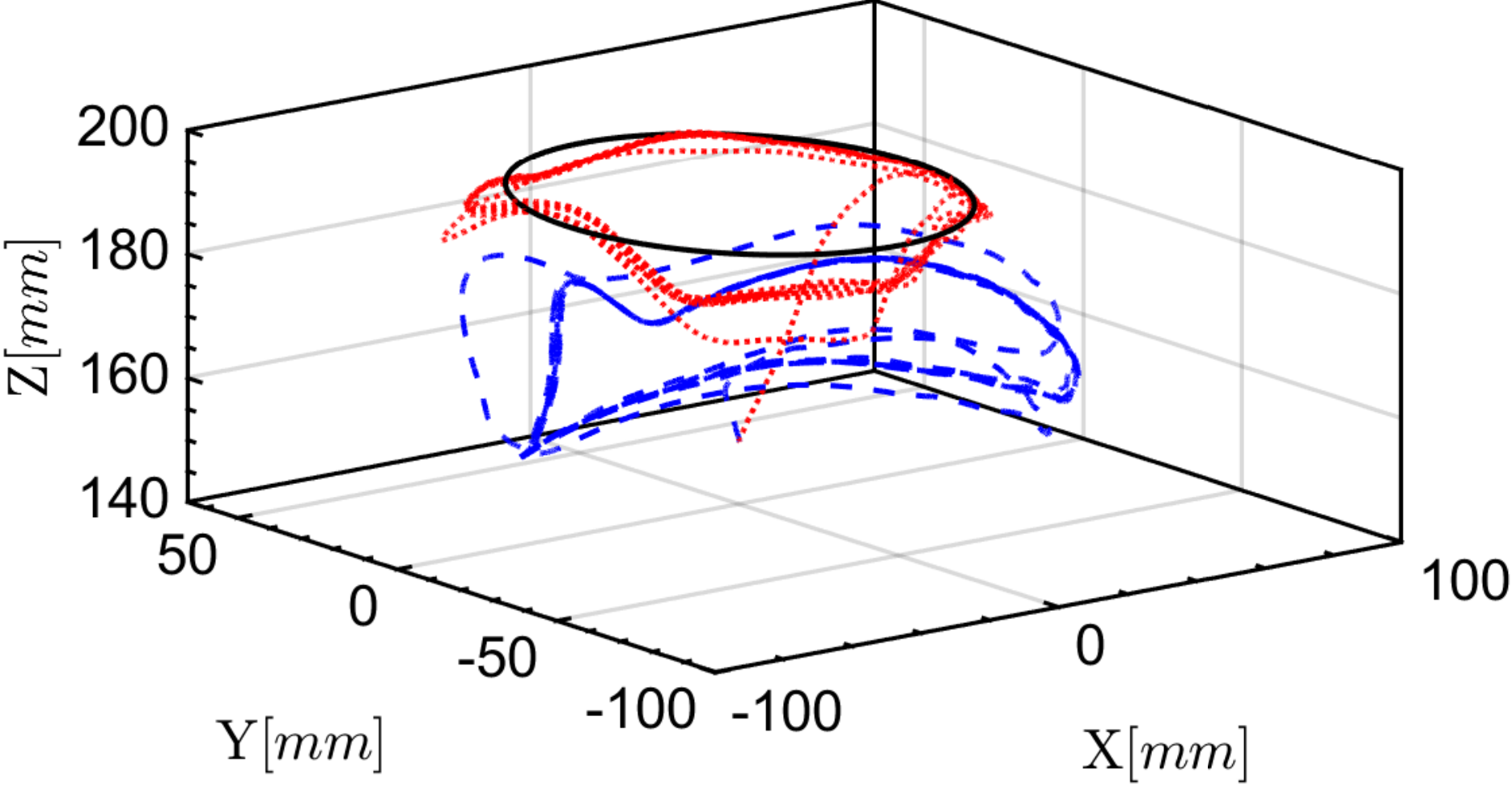}

     \end{subfigure}
     \caption{Trajectory tracking in actuator space and task space. The AP is in red, PDFL in blue and the desired trajectory in black with $\omega = 3 \ rad/s$. 200 g payload has been attached to the system as an external load and disturbance.}
     \label{fig:PathTrack3radLoad200g}
\end{figure}

To further explore the capability of the AP controller, the speed of the desired path has increased to $\omega = 3 \ rad/s$. The results are shown in Fig. \ref{fig:PathTrack3radLoad200g}.
As can be seen, similar to free tracking results in \ref{sub:3radfree}, as the speed increases, the error bound also increases. However, the AP control still retained its stability. 

\subsubsection{Trajectory tracking with the speed of $1 \ rad/s$ subjected to a 500 $g$ payload and external disturbances}
\begin{figure}[t]
     \centering
     \begin{subfigure}[c]{0.15\textwidth}
          \caption{$l_1$ Tracking}
     \includegraphics[width=\textwidth]{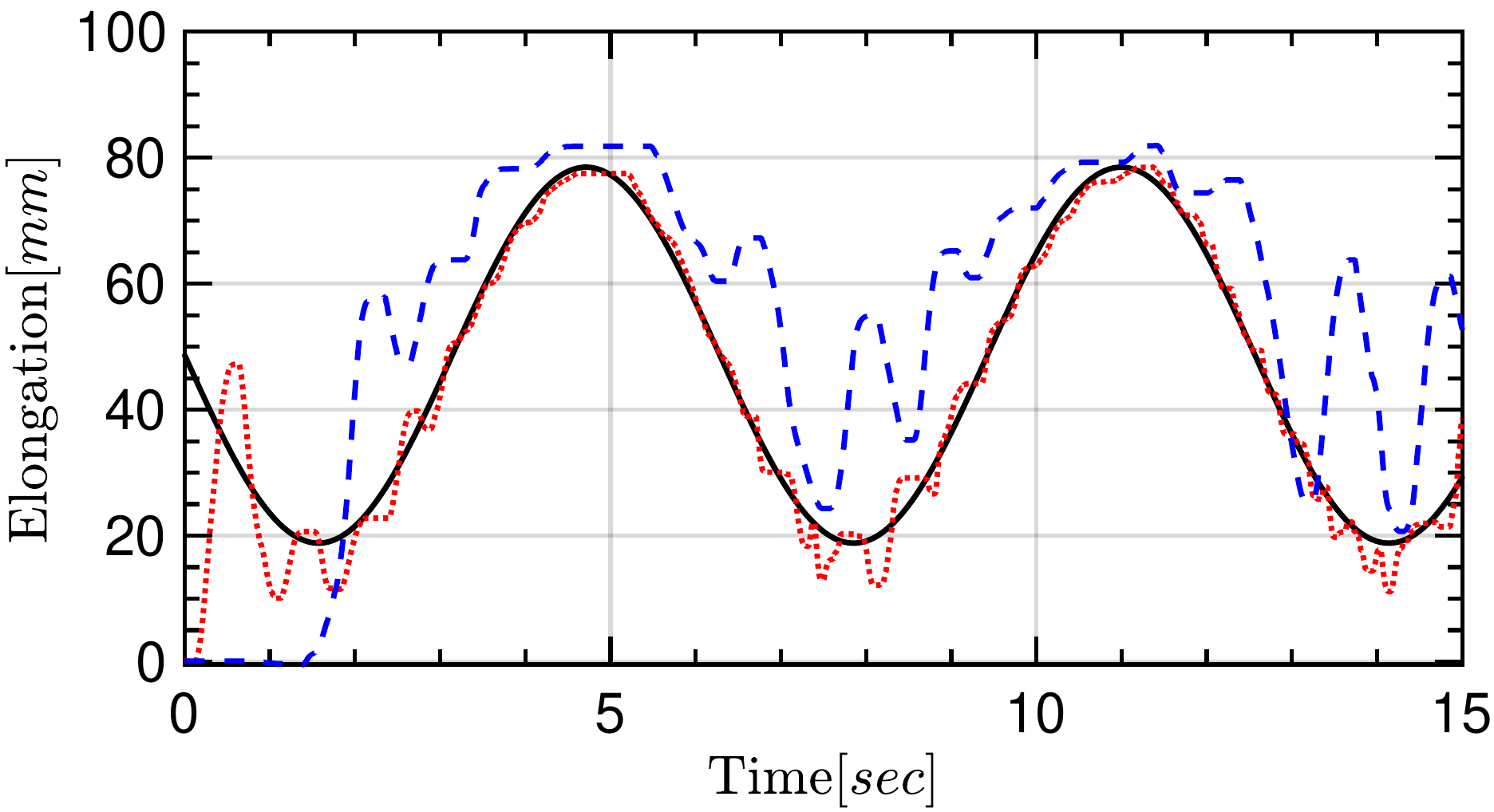}

     \end{subfigure}
     \begin{subfigure}[c]{0.15\textwidth}
          \caption{$l_2$ Tracking}
     \includegraphics[width=\textwidth]{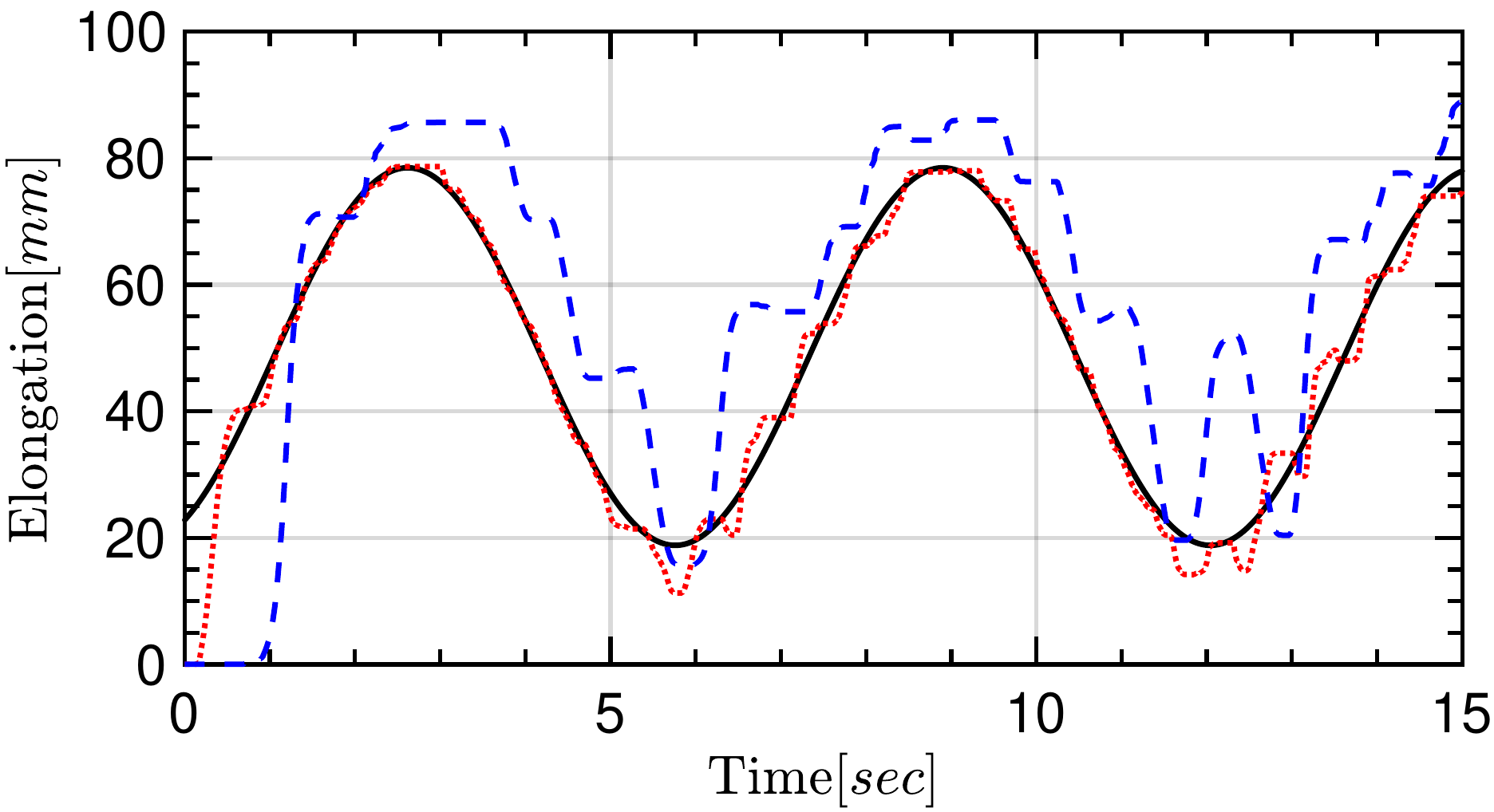}

     \end{subfigure}
     \begin{subfigure}[c]{0.15\textwidth}
          \caption{$l_3$ Tracking}
     \includegraphics[width=\textwidth]{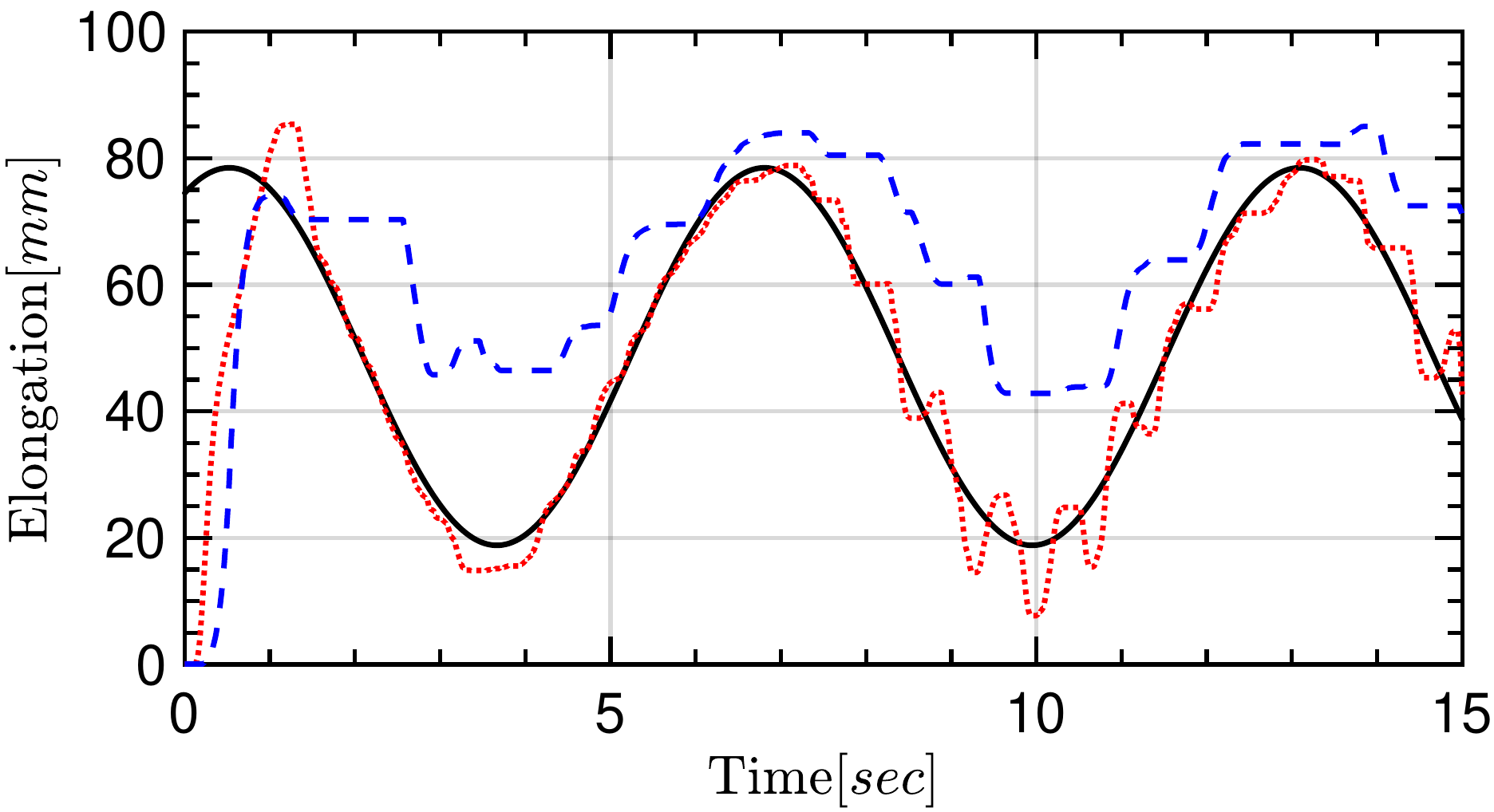}

     \end{subfigure}
     \begin{subfigure}[c]{0.24\textwidth}
          \caption{Position tracking in XY plane}
     \includegraphics[width=\textwidth]{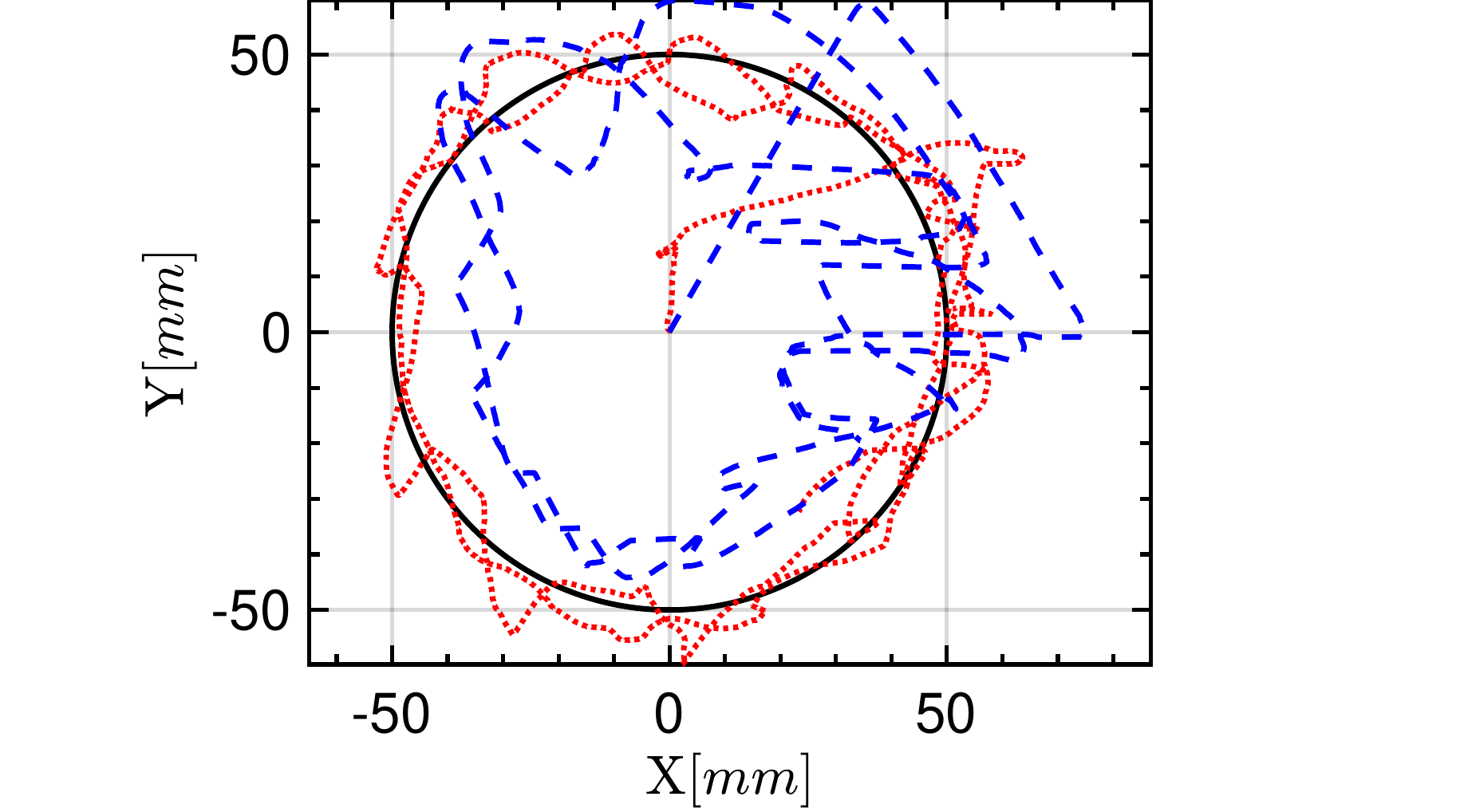}

     \end{subfigure}
     \begin{subfigure}[c]{0.24\textwidth}
          \caption{Position tracking in XYZ plane}
     \includegraphics[width=\textwidth]{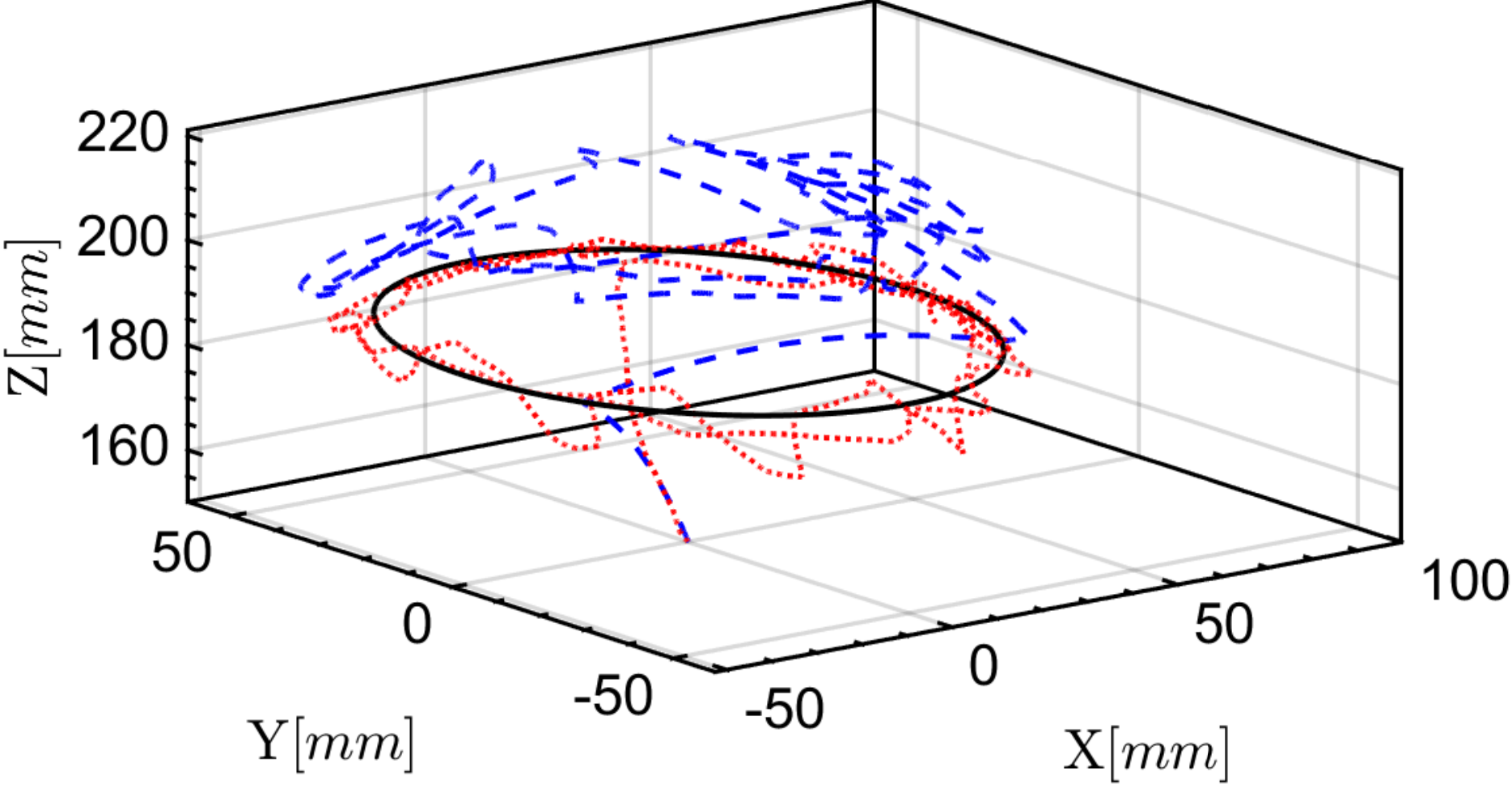}

     \end{subfigure}
     \caption{Trajectory tracking in actuator space and task space. The AP is in red, PDFL in blue and the desired trajectory in black with $\omega = 1 \ rad/s$. 500 g payload has been attached to the system as an external load and disturbance}
     \label{fig:PathTrack1radLoad500g}
\end{figure}

To push the controller even further, we increase the $200 \ g$ load to $500 \ g$. As can be seen in Fig. \ref{fig:PathTrack1radLoad500g}, the disturbance effect is substantial to the system. However, even in this situation, the new design of the robot along with the AP controller maintained the stability, despite the tracking error is significantly increased. 

\subsubsection{Trajectory tracking with the speed of $3 \ rad/s$ subjected to a 500 $g$ payload and external disturbances} 
\begin{figure}[t]
     \centering
     \begin{subfigure}[c]{0.15\textwidth}
          \caption{$l_1$ Tracking}
     \includegraphics[width=\textwidth]{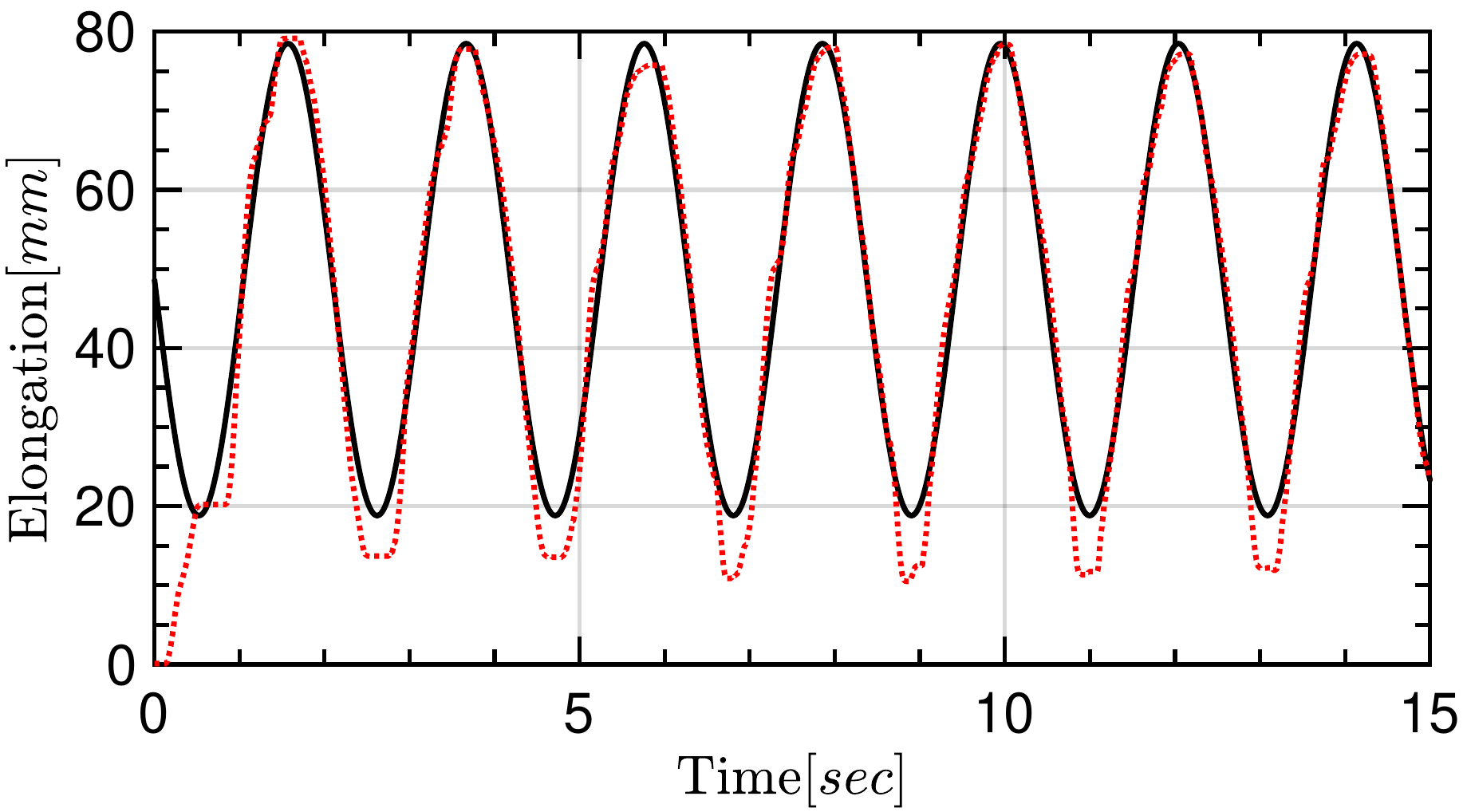}

     \end{subfigure}
     \begin{subfigure}[c]{0.15\textwidth}
          \caption{$l_2$ Tracking}
     \includegraphics[width=\textwidth]{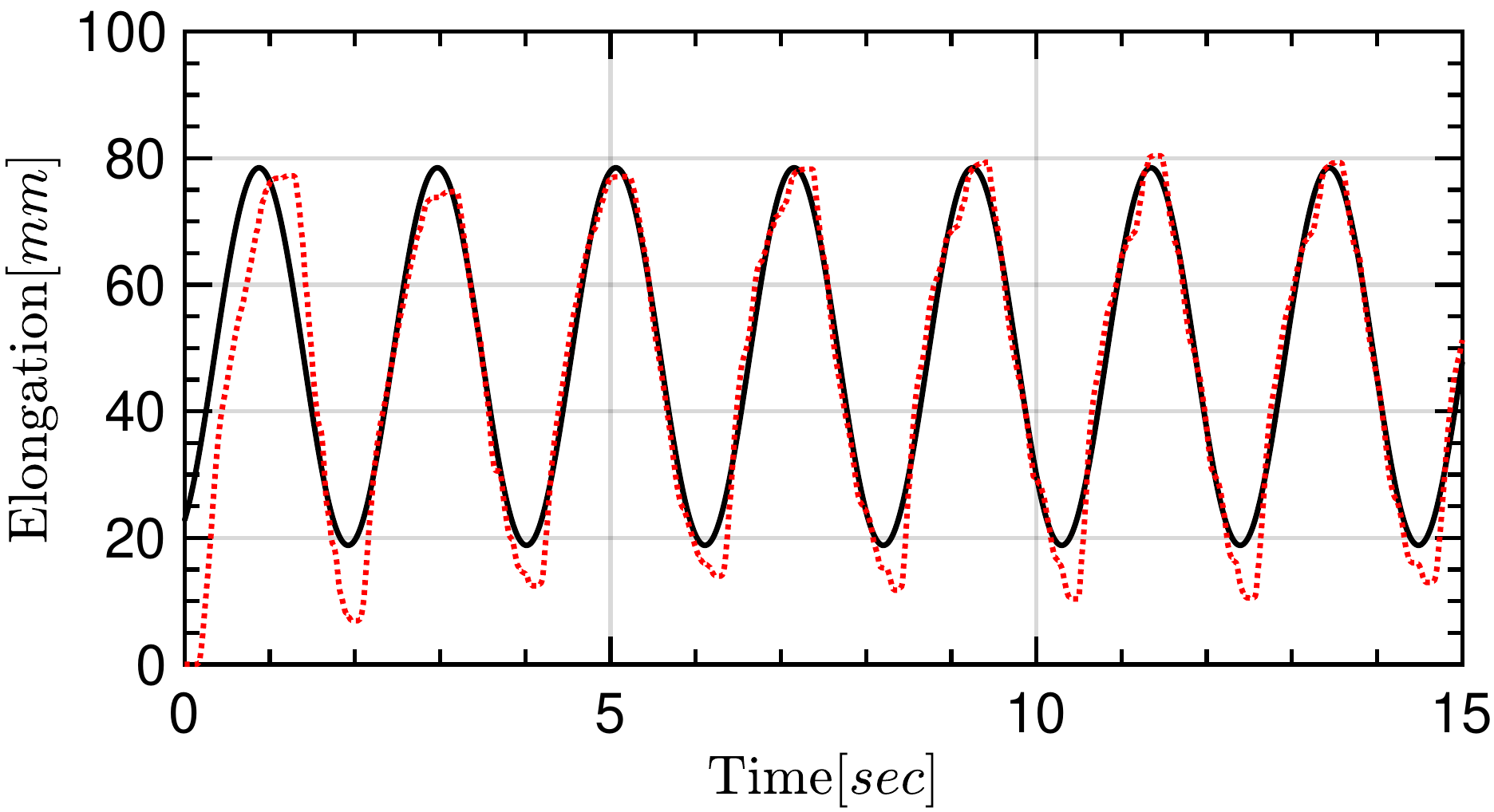}

     \end{subfigure}
     \begin{subfigure}[c]{0.15\textwidth}
          \caption{$l_3$ Tracking}
     \includegraphics[width=\textwidth]{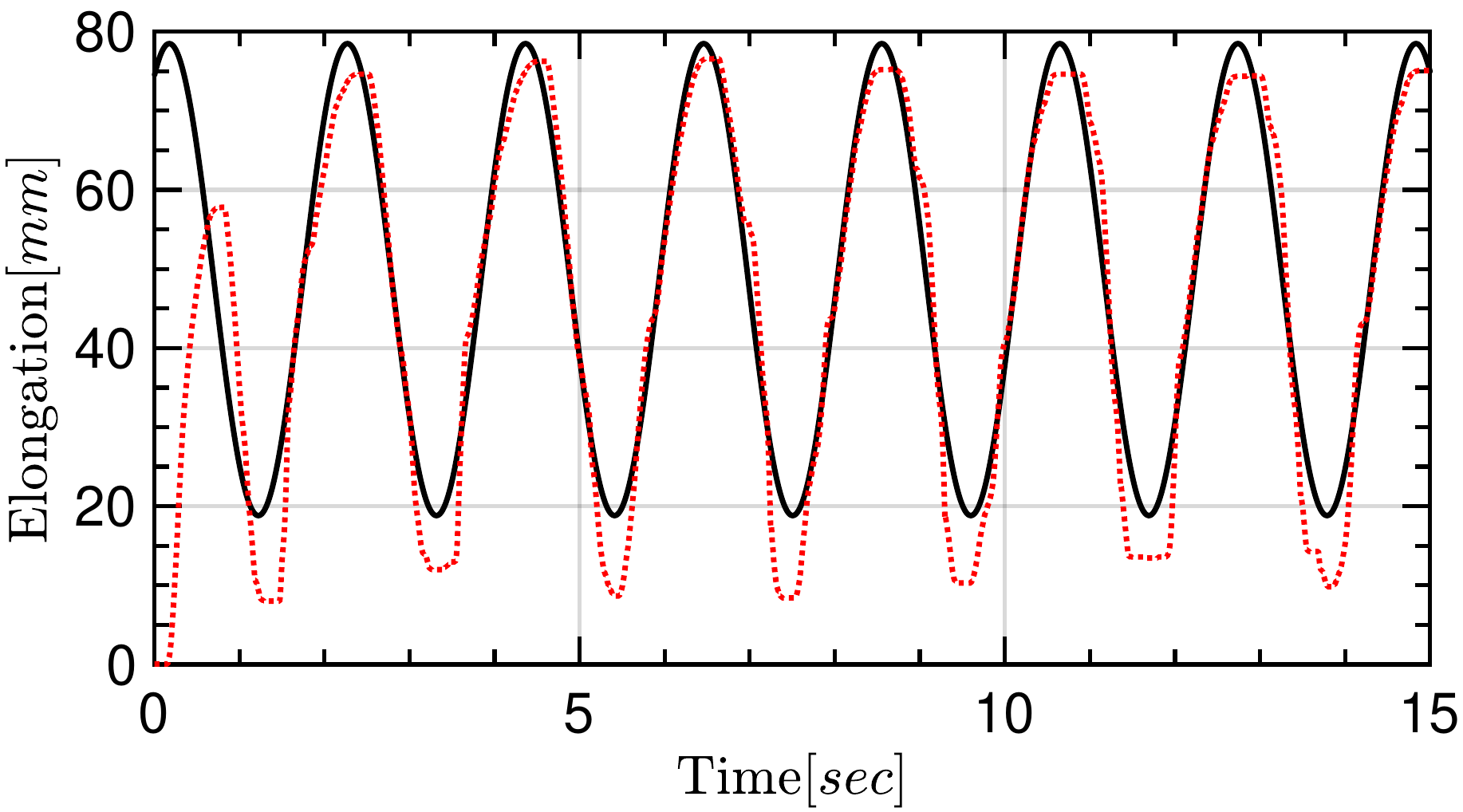}

     \end{subfigure}
     \begin{subfigure}[c]{0.24\textwidth}
          \caption{Position tracking in XY plane}
     \includegraphics[width=\textwidth]{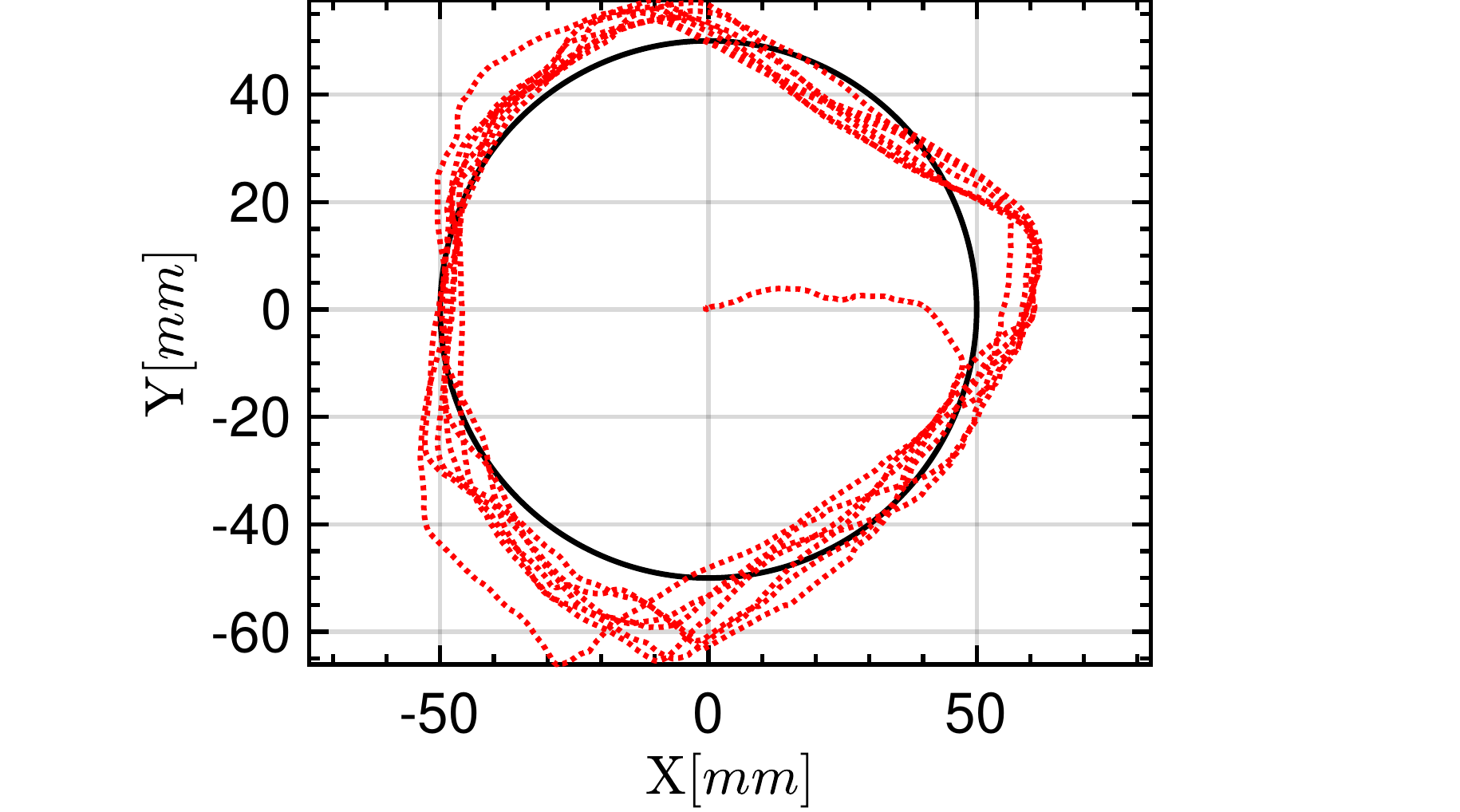}

     \end{subfigure}
     \begin{subfigure}[c]{0.24\textwidth}
          \caption{Position tracking in XYZ plane}
     \includegraphics[width=\textwidth]{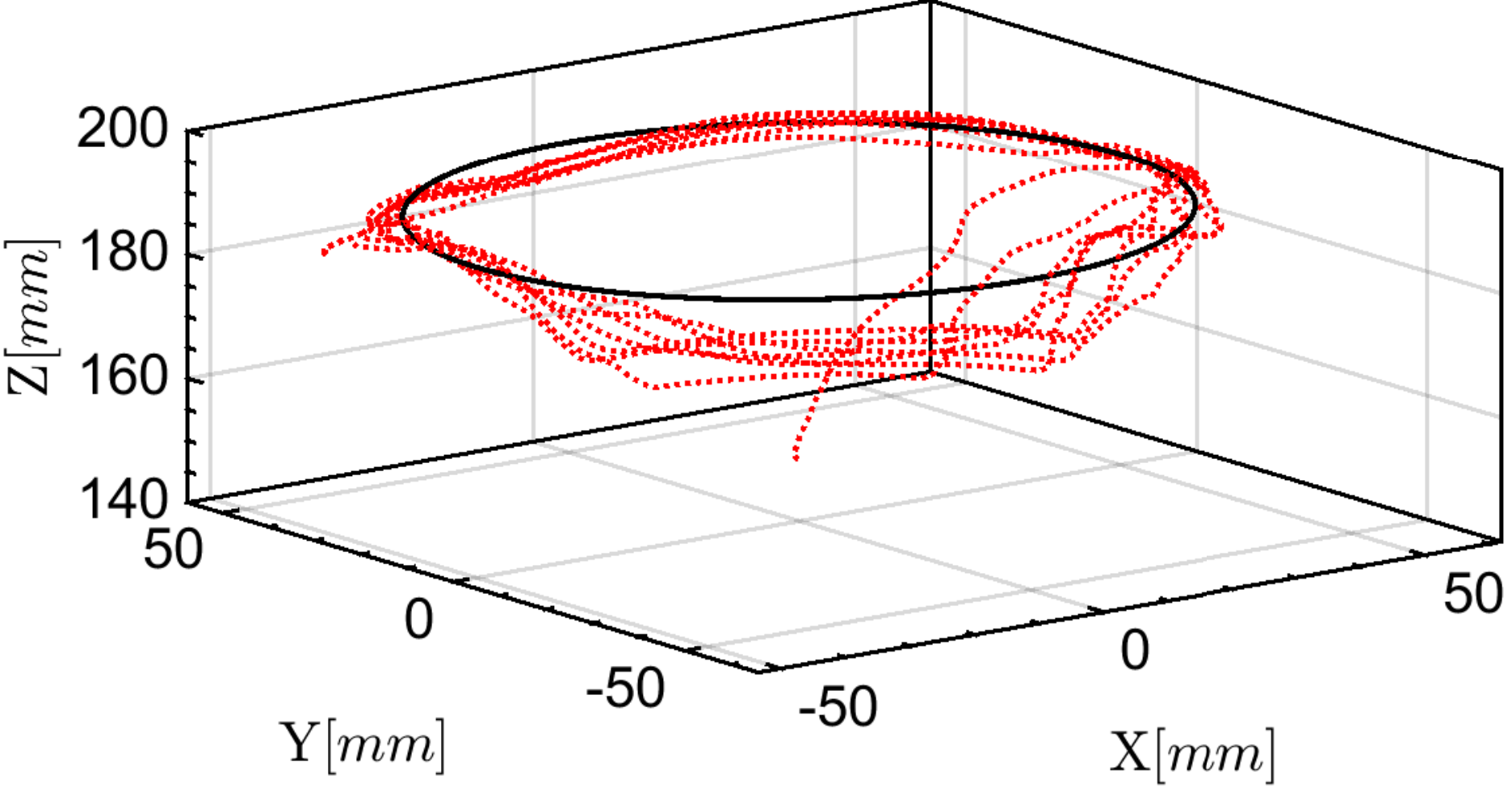}

     \end{subfigure}
     \caption{Trajectory tracking in actuator space and task space. The AP is in red, and the desired trajectory in black with $\omega = 3 \ rad/s$. 500 g payload has been attached to the system as an external load and disturbance}
     \label{fig:PathTrack3radLoad500g}
\end{figure}

In this part, the system is required to follow a 3 $rad/sec$ path with a loosely attached 500 $g$ payload which can be considered as the most challenging experiment for the robot in this study. Note that the PDFL becomes unstable and thus the data was not shown here. As can be seen in Fig. \ref{fig:PathTrack3radLoad500g}, the AP control is able to track the desired signal.

\vspace{-3 mm}
\subsection{Discussion}

To provide the comprehensive overview of the proposed controller in different operation scenarios, we summarize the L2 norm of the error in both actuator-space and task-space (cartesian space) in Fig. \ref{fig:BoxPlot}. Both controller performances during the aforementioned  operation scenarios are included except for PDFL in scenario 6 due to the system instability. 
 
\subsubsection{Effect of path speed on control performance}

\begin{figure}[t]
     \centering
     \includegraphics[width=0.43\textwidth]{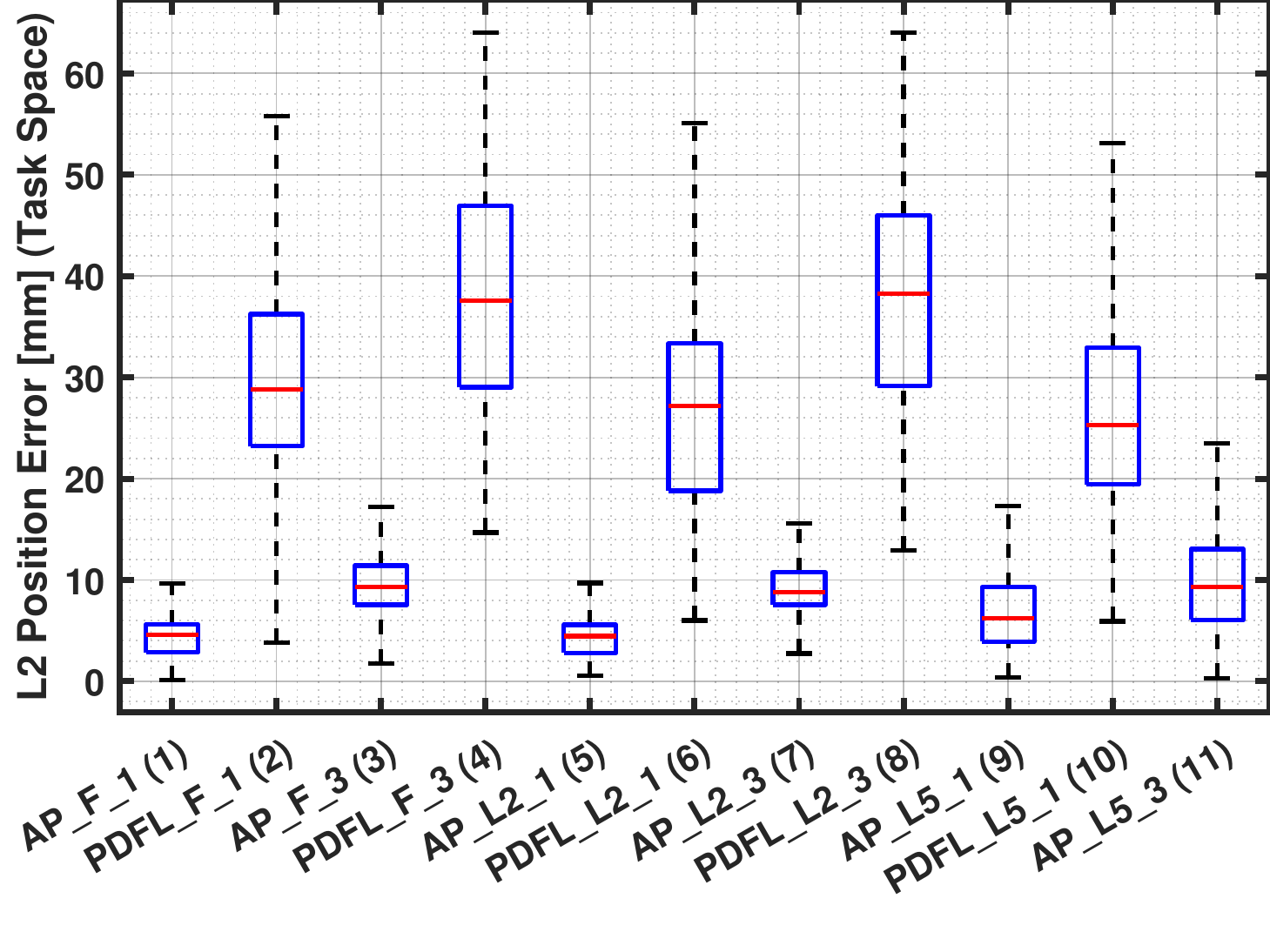}
     \caption{Boxplot for highlighting the results in different scenarios for both AP and PDFL. The code for the X-Axis from left to right is as follows: [Control Type]-[Free(F) or Loaded (L)][m*100 g payload]-[Speed of the path in rad/s] [Data ID]. }
     \label{fig:BoxPlot}
\end{figure}

Based on Fig. \ref{fig:BoxPlot}, it can be seen that the speed will result in the reduced tracking accuracy both in AP and PDFL. This can be further highlighted by comparing the mean of the L2 norm error according to Table \ref{tab:tab2}. It is obvious that  the AP maintained its stability and performance despite the increase of the error, which highlights the robustness and adaptation capabilities. For manipulation purposes,  accuracy and consistency  are  the most important factors, and can be achieved using our proposed method. It can be argued that tracking accuracy in AP can be potentially improved with high gain, but it will inevitably increase the chattering and non-smooth behavior of robot, leading the system to instability. % the control needs to get faster, which can be easily achieved by increasing the gain of the controllers.
% However, this action will result in non-smooth behavior from the robot which will not be acceptable for handling objects. 
The proposed strategies for compensating high-speed  effects are discussed in section \ref{sec:Conclusion}. 

\subsubsection{Disturbance/payload effect on tracking performance}

According to Table  \ref{tab:tab2}, the mean $L_2$ error for PDFL approach remained the same with some slight variation. This indicates that the PDFL algorithm cannot handle the system dynamics or external disturbances. On the other hand, the AP maintained its performance  even with the extreme operation scenario ( 500 g payload and 3 $rad/s$). Despite the increased deviation from the path, it is clear that the system maintained its stability without even changing the control parameters.

\begin{table}[tbh]

    \centering
        \begin{tabular}{|c|c|c|c|c|c|}
        \hline
        \rowcolor{Gray}
             Data ID (Task space) & 1 & 2 & 3 & 4 & 5 \\
        \hline
            Mean [mm] & 6.81 & 30.58 & 11.01 & 38.11 & 6.72 \\
        \hline
        \rowcolor{Gray}
             6 & 7 & 8 & 9 & 10 & 11\\
        \hline
            27.28 & 10.65 & 38.06 & 8.51 & 26.17 & 11.28\\
        \hline
        \end{tabular}
    \caption{Mean L2 error of tracking in task space [mm]. The data ID  is defined in Fig. \ref{fig:BoxPlot} caption.}
    \label{tab:tab2}
\end{table}

\section{Conclusion}
\label{sec:Conclusion}
In this paper, a new reinforced soft robotic arm has been designed to investigate the control strategy during different operation scenarios.  A custom-designed string encoder was introduced for actuator-space measurement, and closed loop control. 
The cost-efficient rotary encoder eliminated the need for high-speed tracking devices and also  removed the need for observer design and sigma modification in the control design \cite{azizkhani2022dynamic} for mitigating the measurement noise effect in the control output
With the robust hardware prototype, we have shown that the proposed AP control scheme can maintain its tracking performance in the challenging operation scenarios where a classic PDFL fails,  which experimentally validated our simulation results  in \cite{azizkhani2022dynamic}. 

In our future studies, we are planning to address the dynamic control of a multi-section soft robotic arm by designing a controller which can adapt to uncertainties and the probabilistic behavior of the robot. Implementing a probabilistic method along with adaptive ability could compensate the uncertainties in a fast and smooth manner, which can result in a more robust and consistent performance at higher speeds that can be crucial for certain industrial applications such as pick-place tasks.

% or


\section*{Acknowledgment}

The authors would like to thank Nyah M. Ebanks for assisting in the development of string rotary encoder.

\ifCLASSOPTIONcaptionsoff
  \newpage
\fi



%\bibliographystyle{IEEEtran}
\vspace{-3 mm}
\bibliographystyle{ieeetr}
\bibliography{references}



%\newpage


%\vfill


\end{document}